\documentclass[pdflatex, sn-mathphys, dvipsnames]{sn-jnl}

\usepackage{booktabs}
\usepackage{siunitx}
\usepackage[inline,shortlabels]{enumitem}
\usepackage{pifont} 
\usepackage{adjustbox}
\usepackage{fancyvrb}
\usepackage{tabularx}
\usepackage{longtable}
\usepackage{bookmark}
\usepackage{pdflscape} 

\newcolumntype{Q}[2]{%
    >{\adjustbox{angle=#1,lap=\width-(#2)}\bgroup}%
    l%
    <{\egroup}%
}
\newcommand*\rot{\multicolumn{1}{Q{50}{1em}}}

\graphicspath{{./figures/}}

\usepackage{hyphenat}  
\usepackage[normalem]{ulem}  

\usepackage{microtype}

\DeclareMathOperator{\softmax}{softmax}
\DeclareMathOperator{\sigmoid}{sigmoid}

\newcommand{\figref}[1]{Fig.~\ref{#1}}    
\newcommand{\Tabref}[1]{Table~\ref{#1}}

\newcommand{\secref}[1]{Section~\ref{#1}}
\newcommand{\equref}[1]{Eq.~(\ref{#1})}
\newcommand{\equsref}[2]{Eqs.~(\ref{#1})--(\ref{#2})}
\newcommand{\appref}[1]{Appendix~\ref{#1}}

\newcommand{\eg}{e.g., }
\newcommand{\cf}{cf.\ }
\newcommand{\ie}{i.e., }
\newcommand{\vs}{vs.\ }
\newcommand{\etal}{\textrm{et al.\ }}

\makeatletter
\def\Url@twoslashes{\mathchar`\/\@ifnextchar/{\kern-.1em}{}}
\g@addto@macro\UrlSpecials{\do\/{\Url@twoslashes}}
\makeatother

\makeatletter
\newcommand*{\Rom}[1]{\expandafter\@slowromancap\romannumeral #1@}
\makeatother
\newcommand{\task}[1]{Task~\Rom{#1}}

\newcommand{\formatact}[1]{\textsf{#1}}
\newcommand{\formatintent}[1]{\textsf{#1}}
\newcommand{\formatid}[1]{\texttt{#1}}

\newcommand{\cmark}{\ding{51}}%

\newcommand{\textitblue}[1]{\textit{\textcolor{blue}{#1}}}

\jyear{2022}%

\theoremstyle{thmstyleone}%
\theoremstyle{thmstyletwo}%

\theoremstyle{thmstylethree}%

\begin{document}

\title[]{{C}ook{D}ial: A dataset for task-oriented dialogs grounded in procedural documents}


\author*[]{\fnm{Yiwei}~\sur{Jiang}}\email{yiwei.jiang@ugent.be}

\author[]{\fnm{Klim}~\sur{\nohyphens{Zaporojets}}}\email{klim.zaporojets@ugent.be}

\author[]{\fnm{Johannes}~\sur{Deleu}}\email{johannes.deleu@ugent.be}
\author[]{\fnm{Thomas}~\sur{Demeester}}\email{thomas.demeester@ugent.be}
\author[]{\fnm{Chris}~\sur{Develder}}\email{chris.develder@ugent.be}

\affil[]{
    \orgdiv{IDLab}, \orgname{Ghent University} -- \orgname{imec}, 
    \orgaddress{%
        \street{Technologiepark~Zwijnaarde~126}, \postcode{9052} \city{Ghent},
        \country{Belgium}%
    }%
}

\abstract{
    This work presents a new dialog dataset, CookDial, that facilitates research on task-oriented dialog systems with procedural knowledge understanding.
    The corpus contains 260 human-to-human task-oriented dialogs in which an agent, given a recipe document, guides the user to cook a dish.
    Dialogs in CookDial exhibit two unique features:
    \begin{enumerate*}[(i)]
    \item procedural alignment between the dialog flow and supporting document;
    \item complex agent decision-making that involves segmenting long sentences, paraphrasing hard instructions and resolving coreference in the dialog context.
    \end{enumerate*}
    In addition, we identify three challenging (sub)tasks in the assumed task-oriented dialog system:
    \begin{enumerate*}[(1)]
    \item User Question Understanding, 
    \item Agent Action Frame Prediction, and
    \item Agent Response Generation.
    \end{enumerate*}
    For each of these tasks, we develop a neural baseline model, which we evaluate on the CookDial dataset.
    We publicly release the CookDial dataset, comprising rich annotations of both dialogs and recipe documents, to stimulate further research on domain-specific document-grounded dialog systems.  
}




\maketitle

\section{Introduction}
\label{sec:intro}

  The last decade has seen a surge of work dedicated to building conversational agents (CA) via 
annual challenges (\eg Dialog System Technology Challenges \citep{gunasekara2020overview-dstc9}) 
or benchmark datasets (\eg WoZ~2.0 \citep{wen-etal-2017-network}, MultiWoZ \citep{Budzianowski2018MultiWOZA}, SGD \citep{rastogi2020towards}).
  To provide meaningful responses, such conversational agents typically rely on some form of background knowledge.
  In  question answering (QA), that knowledge often takes the form of descriptive texts from which an answer is distilled (\eg SQuAD \citep{rajpurkar-etal-2018-know}), 
while in more complicated settings (possibly involving multi-hop reasoning), the system reasons over knowledge graphs (\eg KdConv \citep{zhou-etal-2020-kdconv}).
  Thus, these question answering works are typically  document-grounded.
  Recent QA works further consider more advanced settings involving multi-turn conversations on a given topic rather than single questions (\eg CoQA \citep{reddy2019coqa} and QuAC \citep{choi-etal-2018-quac}), 
and/or span various domains (\eg DoQA \citep{campos-etal-2020-doqa}.
). 
  Logical extensions of this line of work pertain to dialogs including follow-up questions (\eg ShARC \cite{saeidi-etal-2018-interpretation}), 
which may also be initiated by the agent itself and require the system to track both the dialog and the document context (\eg doc2dial \cite{feng-etal-2020-doc2dial}).
  Going beyond answering user questions,  task-oriented dialog systems assist with, \eg flight, hotel and restaurant reservations, and similarly are grounded in supporting knowledge.
  The latter typically takes the form of a well-structured  database, from which the system retrieves information relevant to the user's task at hand (\eg WoZ~2.0 \citep{wen-etal-2017-network}).

  Despite this growing body of work, we note that an under-explored use case in task-oriented systems is that of assisting the user to follow a given procedure.
  A notable exception is the work of Raghu \etal \cite{raghu2021endtoend}, which is based on flowcharts describing the procedure.
  To facilitate further research on procedural assistance dialog systems, we focus on establishing an exemplary dialog dataset and baseline solutions for a conversational agent grounded in textual procedural documents. 
  Thus, our work relates to aforementioned  document-grounded systems. Yet, whereas in these state-of-the-art works the documents contain a descriptive text on a particular subject, in our case those documents describe specific instructions.
  The procedural task dialogs we consider thus concern entities that are manipulated with certain tools, and steps that need to be executed in a particular order, etc.

\begin{figure*}[ht]
    \centering
    \includegraphics[width=.9\textwidth]{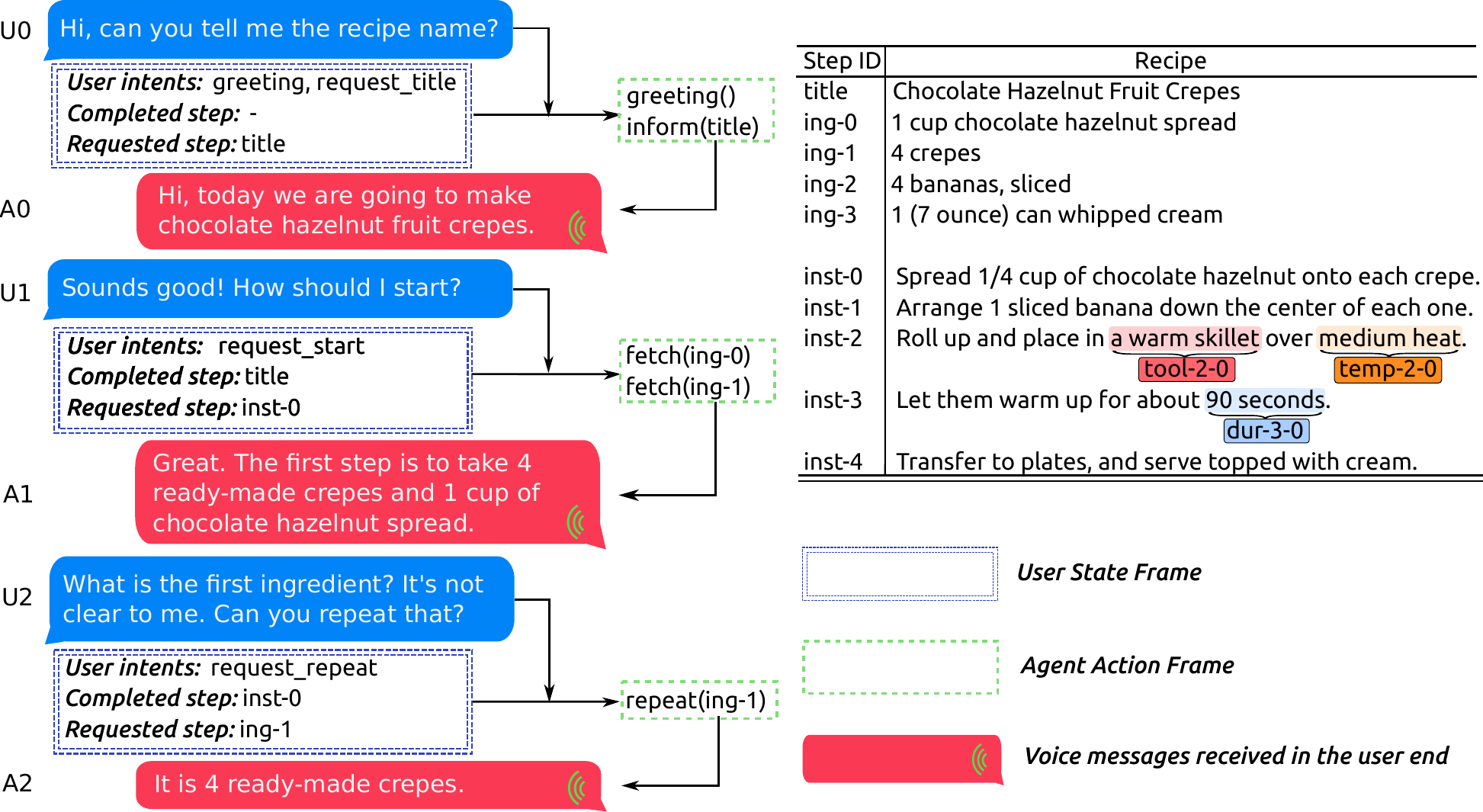} 
    \caption[Caption without FN]{The left part shows a cooking dialog sample in which U and A represent the user and agent respectively.
          The upper right table is the corresponding recipe document.
          Identifiers \formatid{ing} and \formatid{inst} denote ``ingredient'' and ``instruction'' respectively.
           These identifiers are listed in the ``Step ID'' column.
          Within the recipe text, only three entities are highlighted (see full recipe annotations in \figref{fig:full_recipe_annotation}). Further, \formatid{temp} and \formatid{dur} are shorthands for ``temperature'' and ``duration''.
          }
    \label{fig:dialog_anno_sample}
\end{figure*}

  Specifically, we introduce a \textbf{cook}ing \textbf{dial}og (CookDial) dataset, comprising conversations where a conversational agent instructs a user to follow a given (textual) recipe.
  We consider this recipe-based dataset as prototypical for procedural assistance conversations.
  The CookDial dialogs thus reflect the required capabilities of such a procedural CA, implying understanding the procedure's structure, and particularly the chronological steps it comprises, 
as well as tracking the user's state in relation to it, \ie which step the user has reached, and relating the entities/tools in user questions to those in the recipe.
  Thus, to build such procedural CA system we basically need to solve three fundamental challenges:
\begin{enumerate*}[(C1)]
    \item \label{it:ca-step1} understanding the procedural documents (\ie recipes),
    \item \label{it:ca-step2} aligning a dynamic conversation with the accompanying document and,
    \item \label{it:ca-step3} implementing a decision-making process for the CA to generate the most appropriate utterance, constrained by the grounded document.
\end{enumerate*}
  For challenge~\ref{it:ca-step1}, we refer to our earlier work on information extraction from procedural texts~\cite{jiang-etal-2020-recipe}.
  In the current paper, we focus on challenges~\ref{it:ca-step2}--\ref{it:ca-step3}.

  To illustrate these challenges, \figref{fig:dialog_anno_sample} lists an exemplary conversation for an annotated crepe recipe shown in the right part.
  Note that we assume the recipe has already been processed~\cite{jiang-etal-2020-recipe} to identify the constituent ingredients and instruction steps (respectively marked with \texttt{ing-}\textit{i} and \texttt{inst-}\textit{j}), 
as well as entities such as tools and attributes (\eg temperature, duration) mentioned within those steps.
  In challenge~\ref{it:ca-step2}, we need to determine the \emph{User State Frame} for each user utterance, i.e,  
to establish which recipe step the user has completed as well as the step which the user is requesting details about.
  This will be {\task{1}}, User Question Understanding.
  Next, the decision process of the agent's response is annotated in an \emph{Agent Action Frame}, phrased as a list of at least one function-like string specifying 
\begin{enumerate*}[(a)]
    \item an \emph{agent act}, and
    \item an \emph{argument pointer} linking to an item of the recipe (\ie an ingredient, a full instruction, or an entity within it).
\end{enumerate*}
  We will split the challenge~\ref{it:ca-step3} into two subtasks:
{\task{2}} will be Agent Action Frame Prediction; \task{3},~ Agent Response Generation, will be 
to generate natural responses given a user question, the dialog context and extra information from the agent action frame.
  For each of these tasks we will provide a corresponding baseline model, based on deep neural networks with pretrained language models.

  In summary, this paper offers three contributions:
\begin{itemize}[topsep=0pt,itemsep=0pt]
    \item We create the CookDial dataset, which to the best of our knowledge is the first that grounds the knowledge of dialog systems in procedural documents.
    The final dataset consists of 260 dialogs and 260 corresponding recipes (\secref{sec:cookdial-dataset}). 
    The dataset and codes are available at \url{https://github.com/YiweiJiang2015/CookDial}.
    
    \item We introduce an annotation scheme (\secref{sec:annotation_details}) for such procedural document-grounded dialog systems, in particular, cooking recipe documents. 
    Specifically, we propose symbolic annotations that assign unique identifiers to steps and entities to facilitate the alignment problem between dialogs and grounding documents. 
    We further propose two annotation phases, separating the user state tracking problem and the agent decision-making process.
    
    \item We identify three challenging tasks (\secref{sec:task_definitions}) and propose baseline models for each (\secref{sec:models_and_results}), which we evaluate on the CookDial dataset: 
  User Question Understanding (\task{1}), Agent Action Frame Prediction (\task{2}) and Agent Response Generation (\task{3}). 
\end{itemize}
  We thus aim to stimulate further research on procedural conversational agent systems, as a subfield of document-grounded dialog models.

  \begin{table}[htpb]
    \small
    \caption{Summary of the characteristics of CookDial compared to other document-grounded dialog datasets.} 
    \label{tab:dataset_feature}
    \begin{tabular}{lcccccc}
                                                         & \rot{CookDial (ours)}  & \rot{doc2dial \citep{feng-etal-2020-doc2dial}}  & \rot{QuAC \citep{choi-etal-2018-quac}}
                                                          & \rot{CoQA \citep{reddy2019coqa}}   & \rot{DoQA \citep{campos-etal-2020-doqa}}   
                                                          & \rot{ShARC \citep{saeidi-etal-2018-interpretation}}  
                                                          \\
        \toprule
        \textbf{Annotation Features} \\
                             Task-oriented dialogs     & \cmark & \cmark &       &        &        & \cmark \\
                TTS setting$^\textrm{\dag}$            & \cmark &        &       &        &        &        \\
    
    Rephrase grounding spans  & \cmark & \cmark &       &        &        &        \\
    
    User intent annotation      & \cmark &        &       &        &        &        \\
                             Agent act annotation        & \cmark &        &       &        &        &        \\
                             Grounding span annotation   & \cmark & \cmark & \cmark& \cmark &        &       \\
                            
        \midrule
        \textbf{System Abilities} \\
                             Procedural understanding    & \cmark & \cmark &       &        &        &       \\
                       User intent identification  & \cmark &        &       &        &        &        \\
                    User question grounding        & \cmark & \cmark &       &        &        &        \\
                             Agent response grounding    & \cmark & \cmark & \cmark& \cmark &        & \cmark \\
                             Document retrieval          &        & \cmark &       &        & \cmark &        \\
        \midrule
        \textbf{Statistics} \\
                             \#\,Dialogs                 & 260       & 4,800     &13,594 & 8,399  & 2,437  & 6,637 \\
                   \#\,Utterances per dialog   & 35        & 14        & 7.2   & 15.2   & NA     & NA \\ 
                             \#\,Grounding documents       & 260       & 480       & 8,854 & 8,399  & 1,908  & 948   \\
        \bottomrule
    \end{tabular}
    \\ \medskip
    \parbox{.8\textwidth}{$^\dag$: TTS, or Text-to-speech, implies that the agent's textual replies are converted to speech messages in the Wizard-of-Oz dialog collection process.; \enspace \textbf{\#}: Number of; \enspace NA: Not available;}
  \end{table}


\section{Related Work}
\label{sec:related}

  As indicated above, we consider a conversational agent (CA) to assist a user to execute a given procedure, as described in a textual document. 
  Thus, our work is a specific use case of a  document-grounded dialog system (DGDS).
  Furthermore, a core capability required by our CA is to grasp the meaning of utterances in the dialog (and relate them to the document). 
  Therefore, our work is also in line with the area of conversational semantic representation.
  Below we sketch related work in both areas, and position our proposed system against it.

\subsection{Document-grounded dialog systems (DGDS)}
  Given that a reasonable fraction of a conversation in our procedural assistant setting amounts to resolving user questions, our work is closely related to the field of conversational question answering.
  In this area, recently various datasets have been released, including QuAC~\citep{choi-etal-2018-quac}, CoQA~\citep{reddy2019coqa},  DoQA~\citep{campos-etal-2020-doqa}
and ShARC~\citep{saeidi-etal-2018-interpretation} .

  CoQA and QuAC have similar settings where a multi-turn conversation is grounded in a text passage (\eg a Wikipedia article). 
  They go beyond reading comprehension challenges like SQuAD~\citep{rajpurkar-etal-2018-know} by requiring the agent to resolve coreferences, ellipsis and contextual reasoning in the dialog history.
  Compared to QuAC and CoQA, we note that DoQA is more oriented towards addressing information retrieval needs, as it is collected from FAQ sites and covers multiple domains. 
  In particular, DoQA includes conversational search tasks \citep{burtsev2017search} that require models to retrieve relevant documents across different domains, given a limited dialog context.
  In ShARC, the information request is further assumed to be underspecified, requiring the CA to pose follow-up questions (based on rules/conditions expressed in the text documents containing the information).
  This implies that a substantial fraction of user utterances are relatively easy to understand (\eg many yes/no answers to the CA's follow-up question), meaning that the system complexity mainly lies in understanding the supporting textual documents (\ie the rules/conditions they express).

  Aforementioned QuAC, CoQA and ShARC datasets require the CA to ``understand'' the textual documents, but mainly on the level of relatively short text spans. 
  The more recent doc2dial \citep{feng-etal-2020-doc2dial} is a document-grounded dialog dataset in which addressing the user question necessitates reasoning across paragraph- and document-level structures.
  Furthermore, rather than pure information-oriented conversations (in the cases of QuAC, CoQA and DoQA), doc2dial is collected in a more goal/task-oriented setting (\eg how to get certain financial benefits).
  This setting makes doc2dial the closest to our work. 
  However, since doc2dial grounds dialogs in documents specifying administrative rules/regulations, the proposed models still lack clear understanding of  step-wise procedures.
  Further, we note that the CA system's ``understanding'' of the supporting textual documents in above datasets closely follows that in machine reading comprehension works~\citep{rajpurkar-etal-2018-know}, by focusing on extracting text spans as answers from documents. 
  However, they show less interest in pragmatic properties of dialogs such as dialog acts, which are important in the semantic understanding of dialogs.
  Our work goes further by providing rich annotations of dialog acts, based on which we introduce semantic representations, \ie user state frames and agent action frames that abstract away from the literal forms (we refer to \secref{sec:conversational_semantic_repr} for further discussion).

  To summarize the characteristics of above DGDS datasets, and how our newly created CookDial compares to them, \Tabref{tab:dataset_feature} outlines their characteristics in terms of
\begin{enumerate*}[(i)]
\item \label{it:dgds-features} annotation features,
\item \label{it:dgds-abilities} system abilities, and
\item \label{it:dgds-stats} basic statistics.
\end{enumerate*}
  The unique features of our CookDial dataset are twofold.
  First, we use text-to-speech (TTS) in the collection process of the dataset, which impacts the dialog flow in that it induces user questions for clarification/repetition.
  Our motivation for assuming a voice interface is that not having to read the textual instructions avoids distraction from executing the procedural actions.
  Second, we also include agent act annotations: similar to the user intent, we also include a formal, functional notation of the CA response (\cf the Agent Action Frame in \figref{fig:dialog_anno_sample}).
  This action frame consists of (a list of) function(s) that indicate the semantics to express in the response, with arguments grounded in the document (\eg entities or steps from the recipe).
  Further, as in doc2dial, the agent annotator is allowed to paraphrase entities/statements of the grounding spans in the documents supporting the conversation (as opposed to purely copying document text spans).

  In terms of system abilities, particularly the procedural understanding is both unique and crucial in our procedural CA setting.
  It implies that the CA needs to be able to clearly identify the distinct steps and their order in the procedure.
  The only other dataset that to some extent implies some ``procedural'' notions is doc2dial,  which however pertains to identifying certain rules or conditions (\eg requirements that need to be fulfilled to warrant benefit eligibility) rather than an ordered sequence of distinct actions.
  Finally, from the statistics it is clear that the size of CookDial is much more modest than others, largely because of time and budget constraints for our data collection.
  Still, CookDial covers all features but the document retrieval ability and has been carefully designed for building a DGDS with procedural knowledge, thus filling a notable gap in the current body of research works.

\subsection{Conversational Semantic Representation}
\label{sec:conversational_semantic_repr}

  The level of ``understanding'' the user utterances in dialog systems gradually evolved in terms of the complexity of meaning representation.
  In specific domain contexts, such as task-oriented CAs to assist users in finding/booking restaurants etc., a slot-filling approach is commonly adopted \citep{Budzianowski2018MultiWOZA, Henderson2014TheTD, RojasBarahona2017ANE, el-asri-etal-2017-frames}:
a predefined template of slots is to be filled by string values.
  To perform the task at hand, these slot values are used to send queries to a database system, whose response then allows the user to make a decision or require further information.
  This sequential slot-filling process stores dialog states in a flat semantic frame, which eases the data collection of annotations, yet significantly constrains the possible interaction among utterances in a complex dialog flow.
  
  Further research in this domain focuses on representing utterances with richer semantic meanings by moving beyond the flat semantic frames.
  The authors of~\citep{kollar-etal-2018-alexa,gupta-etal-2018-semantic-parsing,aghajanyan-etal-2020-conversational} extend the classical intent-slot framework by using nested intents in slots, 
thus allowing for more complex queries that contain several intents and corresponding slot-value pairs.
  These nested structures still rely on the static attributes of the predefined database, comprising, \eg food type, price, restaurant location.
  Dialog systems based on such a static database need to fill in these attributes (slots) during the conversation.
  In fact, the CA cares little about the occurring ordering of the slots, as long as it eventually can form a valid query 
(\eg the user can ask for the price and location in any order as he or she likes, without influencing the query result).
  However, in procedures such as recipes in case of CookDial, the order of the steps is crucial and needs to be strictly followed.
  Further, entities are not static across this procedure: in our CookDial case, ingredients are manipulated and change state, 
implying that the agent may need to be aware of these changes (\ie tracking how the entity evolves throughout the procedure) to answer user queries (as opposed to, \eg restaurant attributes which remain constant).
  Also, these procedural ``actions'', which are formally captured in semantic frames, may have varying numbers and types of arguments, thus implying a greater diversity than aforementioned pre-defined slots in more traditional task-oriented CA cases.
  This nature of procedural documents makes the development of a dialog system based on CookDial quite challenging.


\section{CookDial Dataset Description}
  \label{sec:cookdial-dataset}
  
  Our CookDial dataset contains 260 dialogs, each of which corresponds to one recipe document. 
    To generate dialogs, we employed 3 paid workers to work in pairs, conversing using a web chat platform.
    Workers were asked to switch between the roles of user and agent after each of their conversations. 
    Later, 2 experts performed annotations in two phases:
    \begin{enumerate*}[(i)]
        \item they first annotated all the entity identifiers within each recipe document (see \secref{sec:recipe_annotation} for more details),
        \item then they annotated the user state frames (\secref{sec:user_question_annotation}) and agent action frames (\secref{sec:agent_response_annotation}) for each dialog. 
    \end{enumerate*}
    In total, it took approximately 600 hours to curate CookDial, of which 350 hours ($\sim$117 hours per worker) were spent on generating the dialog conversations ($\sim$40 minutes per dialog), 
  50 hours ($\sim$25 hours per expert) for document annotations and 150 hours ($\sim$75 hours per expert) for dialog annotations, excluding about 50 hours of initial annotations to develop the format and guidelines.
    This time cost reflects the difficulty of collecting human-to-human dialogs, especially in a document-grounded setting.
  
  \subsection{Recipe collection}
  
    We use recipe texts from the RISeC corpus\footnote{\url{https://github.com/YiweiJiang2015/RISeC}} \citep{jiang-etal-2020-recipe},
  since it contains already annotated recipes, where the entities involved (\ie ingredients, tools) as well as the actions on them (semantically represented as relations between predicates and their arguments) already have been rigorously identified.
    In CookDial, we extend the original RISeC annotations by assigning unique identifiers to entities.
    See \secref{sec:recipe_annotation} for more details.

  \subsection{Dialog collection}
    As indicated previously (\secref{sec:related}), a substantial body of recent work has considered document-grounded dialog systems (DGDS), but largely focused on information-oriented conversations, mostly in question answering settings. 
    Conversely, we focus on task-oriented dialogs, specifically focusing on procedural tasks involving a sequence of steps to complete the task, grounded in a document describing these steps textually.
    Therefore, we collected the data with two objectives in mind:
      \begin{enumerate*}[(i)]
      \item grounding dialogs in documents, \ie recipes, and steering the conversation towards a specific goal, \ie complete the cooking process;
      and
      \item collecting high-quality conversations that can be used for building a dialog system or its components.
      \end{enumerate*}
  
    To fulfill the first goal, we adopt the recipe corpus from our previous RiSEC work,
    given its already annotated entities and actions performed on them, allowing linking them further to the conversation utterances in the dialogs.
    In total, we gathered 260 dialogs, using the following procedure.
    
    We adopted the Wizard-of-Oz method in which two workers talked to each other via a web chat platform. 
    Each cooking conversation lasted for 20-40 minutes, depending on the length and difficulty of the recipe.
    We collected exactly one dialog per recipe.
    One of the workers mimicked the `agent', thus having full access to a recipe text, and was responsible for 
  assisting the other annotator, namely the `user', to accomplish the cooking. 
    In the remainder of this section, we will use `agent' and `user' to represent the corresponding workers. 
    The user knew nothing about the recipe except its title.
    To complete the cooking task, the user had to keep asking questions until the recipe was done. (Note that in practice, the user did not actually perform any cooking --- we simply asked the annotators to imagine an actual cooking setting as accurately as possible.)
    
    In our chatting application, the agent's textual input was converted to speech messages (by using a free text-to-speech service\footnote{\url{https://responsivevoice.org/}}) 
  while the user's input remained as is, in textual form.
    Such configuration resembles popular chatbot services like Alexa, Google Assistant and Cortana.
    This sometimes caused the `user' to have difficulty fully understanding utterances from the `agent', either because of text-to-speech quality issues or the complexity of the sentence/word itself.
    This impaired comprehension then drove the `user' to ask questions for clarification or repetition.
    Further, to offer a more understandable answer, the `agent' resorted to segmenting or paraphrasing the original recipe instructions. 
    In this way, the generated dialogs benefited from these clarification questions in terms of enhanced language diversity and non-linearity of dialog flows.
    In contrast, in pilot experiments without the TTS setting, `user'
  annotators were found to lazily ask mostly trivial procedural questions (\eg ``What is the next step?'' or ``What should I do now?''). 
    Such lack of creative questions rendered the dialogs too linear and straightforward, and less representative of dialogs we can expect to encounter in the wild. 
      
  \begin{figure}[h!]
    \centering
    \includegraphics[width=\textwidth]{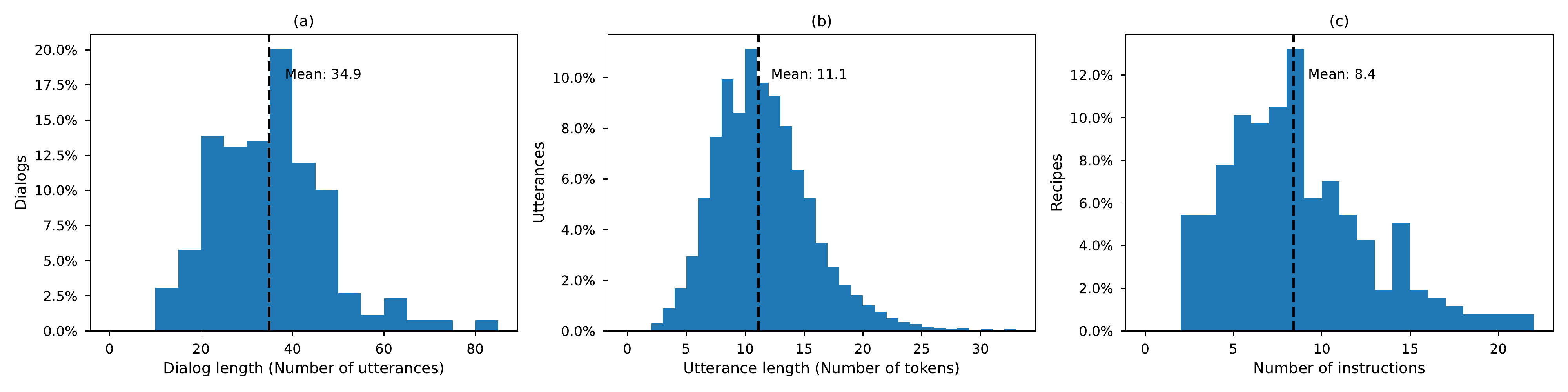}
    \caption[blah]{Statistics of dialogs and recipe documents in CookDial:
      \begin{enumerate*}[(a)]
      \item dialog distribution by the dialog length, 
      \item Utterance distribution by the utterance length,
      \item Recipe distribution by the number of instructions.
      \end{enumerate*}
    }
    \label{fig:statistics_dialog}
  \end{figure}
  
    After the first round of dialog generation, the expert annotators cleaned the data by fixing typos, wrong entities, etc., as marked by annotators after their session was finished or by our expert annotators.
    The final step was to merge the sentences from the same speaker into one utterance,
  as the user or agent sometimes sent a follow-up message to correct typos or mistakes. 
    \figref{fig:statistics_dialog} shows statistics of the 260 dialogs and recipes. 
    The average dialog length is 34.9 utterances. 
    There are 9,068 utterances in total, with an average length of 11.1 tokens per utterance.


  \section{Annotation Details}
  \label{sec:annotation_details}
    This section introduces the annotation schema for our corpus regarding the recipe texts and the dialogs.
    For the recipes, we extend the original RISeC dataset \cite{jiang-etal-2020-recipe} by assigning an identifier to each entity, in order to facilitate the dialog annotation, as described in \secref{sec:recipe_annotation}. 
    For the dialogs, we annotate both the user's and agent's utterances using dialog acts and entity identifiers from the recipe annotations.
  
  \subsection{Recipe annotation}
  \label{sec:recipe_annotation}
    The RISeC corpus \citep{jiang-etal-2020-recipe} contains entity and predicate-argument relation annotations on 260 recipe texts.
    It focuses on revealing semantic structures within the instruction part of recipes, while ignoring the ingredient list 
  that actually provides basic entities (\ie ingredients and their quantities, or their mentions as part of descriptive instructions further down the recipe) for dialogs.
    However, in the context of recipe-grounded dialogs, these entities are key, since user-issued questions are often entity-centric (\eg ``Could you tell me the recipe title?'' and ``What is the next ingredient?'').
    Moreover, the user often asks for a whole instruction from the recipe (\eg ``That is done. What shall I do now?''), which cannot be uniquely identified by the original RISeC annotations.
    
\begin{figure*}[h]
      \centering
      \includegraphics[width=.9\textwidth]{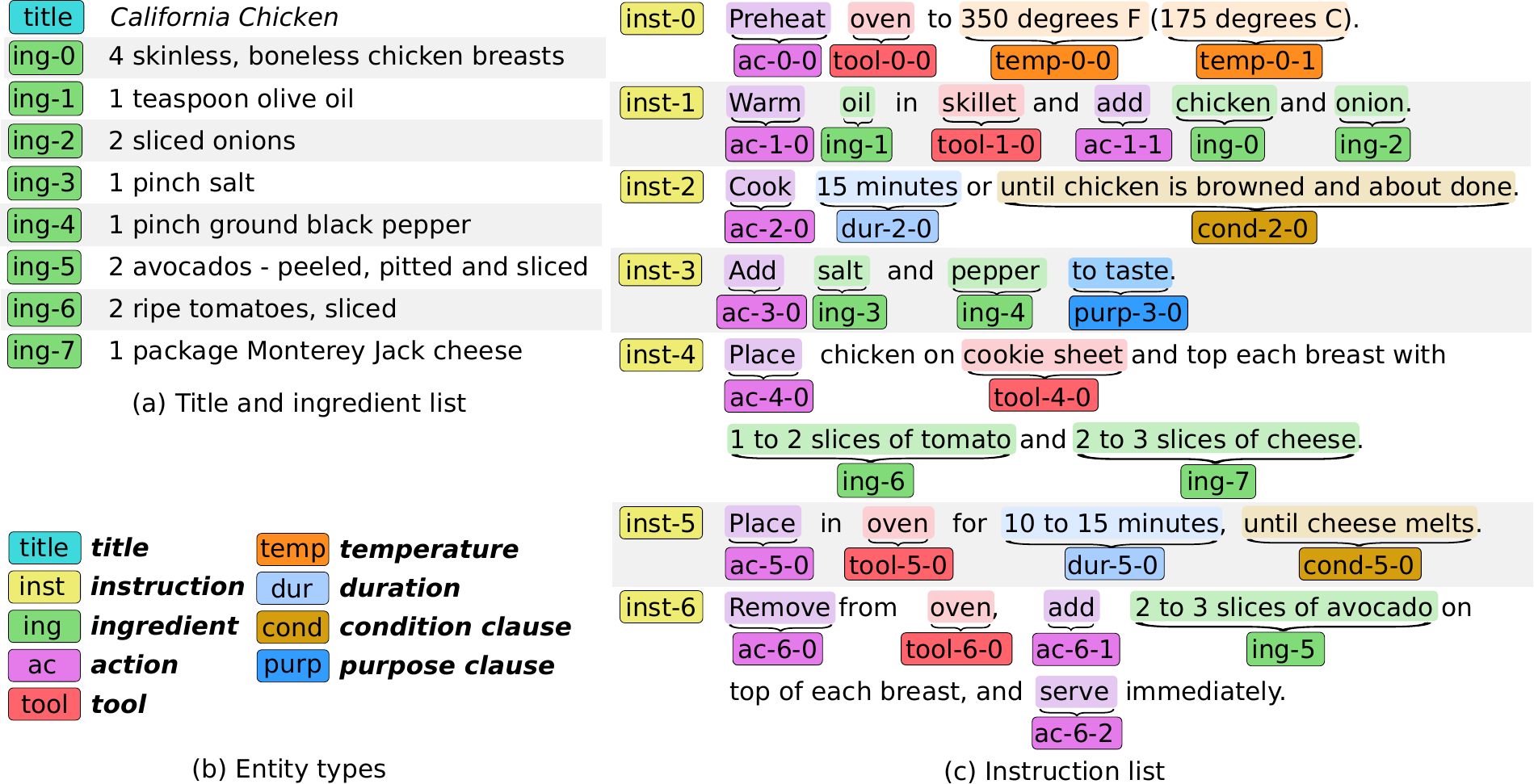}
      \caption{A fully annotated recipe text (viewing in color is recommended).}
      \label{fig:full_recipe_annotation}
    \end{figure*}

    To address these limitations, we extend the original RISeC annotations by assigning unique identifiers to title, ingredients, and instructions, as well as to the entities within the instructions.
    \figref{fig:full_recipe_annotation} shows a fully annotated recipe in which all identifiers are visualized in colored rectangular boxes. 
    The title ``California Chicken'' is tagged as \texttt{title}, followed by the list of ingredients 
  of which each ingredient (typically including an amount and description) is assigned an identifier formatted as \texttt{ing-}\textit{i}, 
  where \textit{i} indicates the rank of the considered ingredient in the list. 
    For example, the last ingredient \texttt{ing-7} in \figref{fig:full_recipe_annotation} corresponds to ``1 package Monterey Jack cheese''. 
    The ingredient list is followed by a list of instructions, each assigned an identifier \texttt{inst-}\textit{j}, where \textit{j} denotes the instruction's index in the ranked list of instructions.
    In the instruction part, all the mentions of ingredients are assigned the corresponding identifier from the ingredient list, even when paraphrased or slightly changed.  For example, the description ``1 to 2 slices of tomato'' is annotated as \texttt{ing-6}, referring to the ingredient originally introduced as ``2 ripe tomatoes, sliced''.
    Note that the RISeC originally 
    introduces a number of other entity types for entities that appear in the instructions, such as \texttt{Action}, \texttt{Tool}, \texttt{Temperature}, etc. \cite{jiang-etal-2020-recipe}, as shown in \figref{fig:full_recipe_annotation}(b). We extend these original annotations into identifiers formatted as 
  \texttt{entity\_type-}\textit{j-k}, in which \textit{j} again refers to the instruction index (\ie the same rank as in the  identifier of the instruction where the entity mention appears), and \textit{k} is the rank of entities of the same type within that instruction.  
    For example, in the first instruction \texttt{inst-0}, there are two temperature expressions, \ie ``350 degrees F'' and ``175 degrees C'', which are assigned labels \texttt{temp-0-0} and \texttt{temp-0-1}, respectively. 
    To wrap up, there are four sets of identifiers within each recipe: title identifier, ingredient identifier, instruction identifier and extended entity identifier. 
    
    In order to facilitate the connection between dialog and recipe annotations in the following sections, 
  and for clearly defining prediction targets in \secref{sec:models_and_results}, we define two supersets of identifiers: 
  \begin{enumerate*}[(i)]
  \item the step identifiers, aggregating title, ingredient, and instruction identifiers, 
  and \item the full set of all recipe identifiers comprising the step identifiers as well as the extended entity identifiers.
  \end{enumerate*}
  
    \begin{figure*}[ht]
      \centering
      \includegraphics[width=0.95\textwidth]{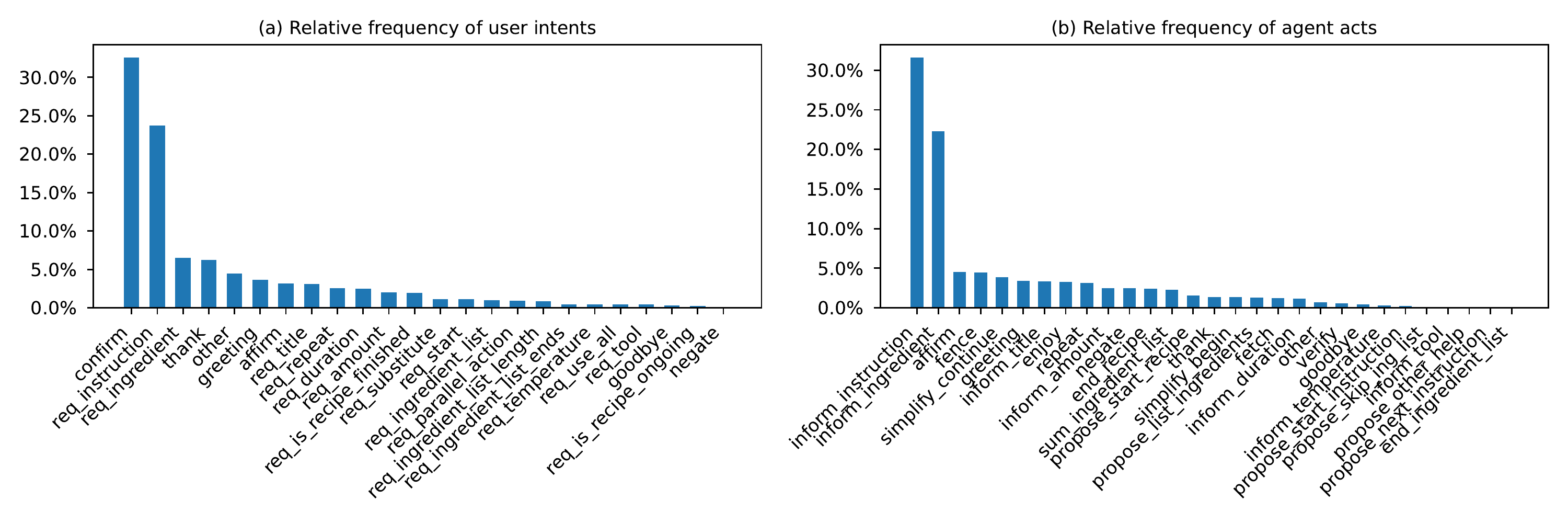}
      \caption{Statistics of user intents and agent acts.}
      \label{fig:freq_intent_agent_acts}
    \end{figure*}

  \subsection{User question annotation}
  \label{sec:user_question_annotation}
  
    Understanding a user's utterance (which we will denote freely as a \textit{question}) usually requires converting a natural utterance 
  into a structured representation, \ie a \emph{User State Frame}. 
    Its annotation task is structured as follows.
  
    The first level of user question annotations identifies one or more suitable user intents. 
    We considered 24 user intents listed in \Tabref{tab:dialog_acts}, some of which are borrowed from the work of \citep{bunt2017dialogue} (\eg \formatintent{greeting}, \formatintent{confirm}, \formatintent{negate}) while others are designed specifically for the cooking dialogs (\eg \formatintent{req\_temperature}, \formatintent{req\_ingredient}). 
  \figref{fig:freq_intent_agent_acts}(a) shows the relative frequency of each intent.
    The top five intents (\formatintent{confirm}, \formatintent{req\_instruction}, \formatintent{req\_ingredient}, \formatintent{thank} and \formatintent{other}) together take up 75\% of the total distribution while the others fall in the long tail region. 
    A detailed description of all these user intents is listed in \Tabref{tab:detailed_user_intent},  \appref{sec:appendix_annotation_details}.
  
    The second level of user question annotations is designed to support user state tracking, and consists of annotating the `completed step' and `requested step' associated with the considered utterance.
    The `completed step' annotation refers to the step the user has just completed or understood, whereas the `requested step' denotes the step most directly following the user request.
    Both are annotated as a step identifier in the recipe (which can be the title, an ingredient, or an instruction ID, as explained in the previous section).
    For example, U1 in \figref{fig:dialog_anno_sample} can be interpreted as the user acknowledging the \formatid{title} (completed step), and asking for the first instruction \formatid{inst-0} (requested step).
  
  \begin{table*}[h!] 
    \caption{dialog acts for the user and agent annotation. \dag: pointers are compulsory.
            \ddag: no argument pointer.} 
      \label{tab:dialog_acts}
      \begin{tabular}{cl}\toprule
                    & Dialog Acts \\
                        \midrule
        User Intents        & greeting, thank, confirm, negate, other, affirm, goodbye \\
                    & req\_start, req\_temperature, req\_instruction, req\_repeat \\
                    & req\_amount, req\_ingredient, req\_use\_all, req\_title \\ 
                    & req\_is\_recipe\_finished, req\_tool, req\_duration \\
                    & req\_is\_recipe\_ongoing, req\_substitute \\ 
                    & req\_ingredient\_list, req\_ingredient\_list\_length \\ 
                    & req\_ingredient\_list\_ends, req\_parallel\_action\\                    
                        \midrule
        Agent Acts\dag      & inform\_instruction, inform\_ingredient, inform\_title \\
                    & inform\_duration, inform\_temperature, inform\_amount \\
                    & inform\_tool, fetch, repeat, verify, simplify\_begin \\ 
                    & simplify\_continue \\
                        \midrule
        Agent Acts\ddag      & greeting, goodbye, affirm, negate, enjoy, fence, end\_recipe \\
              & thank, other, count\_ingredient\_list, propose\_start\_recipe \\
              & end\_ingredient\_list, propose\_list\_ingredients \\ 
              & propose\_skip\_ing\_list, propose\_next\_instruction \\
              & propose\_start\_instruction, propose\_other\_help \\
                        \bottomrule
      \end{tabular}
  \end{table*}
  
  \subsection{Agent response annotation}
  \label{sec:agent_response_annotation}
  
    Besides the user's questions, we also annotate the agent's responses,
  using a different representation: each response is labeled with an action frames consisting 
  of an agent act and, as an argument, an optional pointer to a particular element in the recipe, under the form of an identifier from the full set of recipe identifiers.
    For example, simple phrases like ``hello'' and ``yes'' are annotated as \textsf{greeting}() and \textsf{affirm}() without any argument pointer.
    Taking A1 in \figref{fig:dialog_anno_sample} as another example, we annotate the response ``Great. The first step is to take 4 ready-made crepes and 1 cup of chocolate hazelnut spread.'' as \formatact{fetch}(\texttt{ing-0}); \formatact{fetch}(\formatid{ing-1})''. 
    Pointer annotations, \ie \formatid{ing-0} and \formatid{ing-1}, are essential in this case since the agent act \formatact{fetch} alone cannot grasp the full semantic meaning of this response.
    Notice that the number of agent acts contained in agent action frames varies among different responses: for example, in \figref{fig:dialog_anno_sample}, A0 and A1 have two acts while A2 only has one.
    \Tabref{tab:dialog_acts} lists the 29 agent acts categorized by whether they need an argument pointer or not.
    Again, we borrow some dialog acts from \citep{bunt2017dialogue} (\eg \formatact{affirm}, \formatact{negate})
  while others are invented specifically for the cooking domain (\eg \formatact{propose\_start\_recipe}).
    We label the sentences that do not belong to any act as \formatact{other}.
    Detailed descriptions of the used agent acts are listed in \Tabref{tab:detailed_agent_acts}, \appref{sec:appendix_annotation_details}.
    The relative frequency of agent acts in \figref{fig:freq_intent_agent_acts}(b) shows that \formatact{inform\_instruction} and \formatact{inform\_ingredient}
  are the major acts used by the agent.
  
    As we will validate in \secref{sec:agent_response_generation_task}, not only do the pointer annotations ground the agent answers,
  but they also help to improve the response generation quality in the final system.
    Since the argument pointers rely on the full set of recipe identifiers, 
  the agent action frame annotations are grounded in a more fine-grained level of recipe texts compared to the step trackers in user state frame annotations.

  \subsection{Dialog flow}
  
\begin{figure}[ht]
        \centering
        \includegraphics[width=0.95\textwidth]{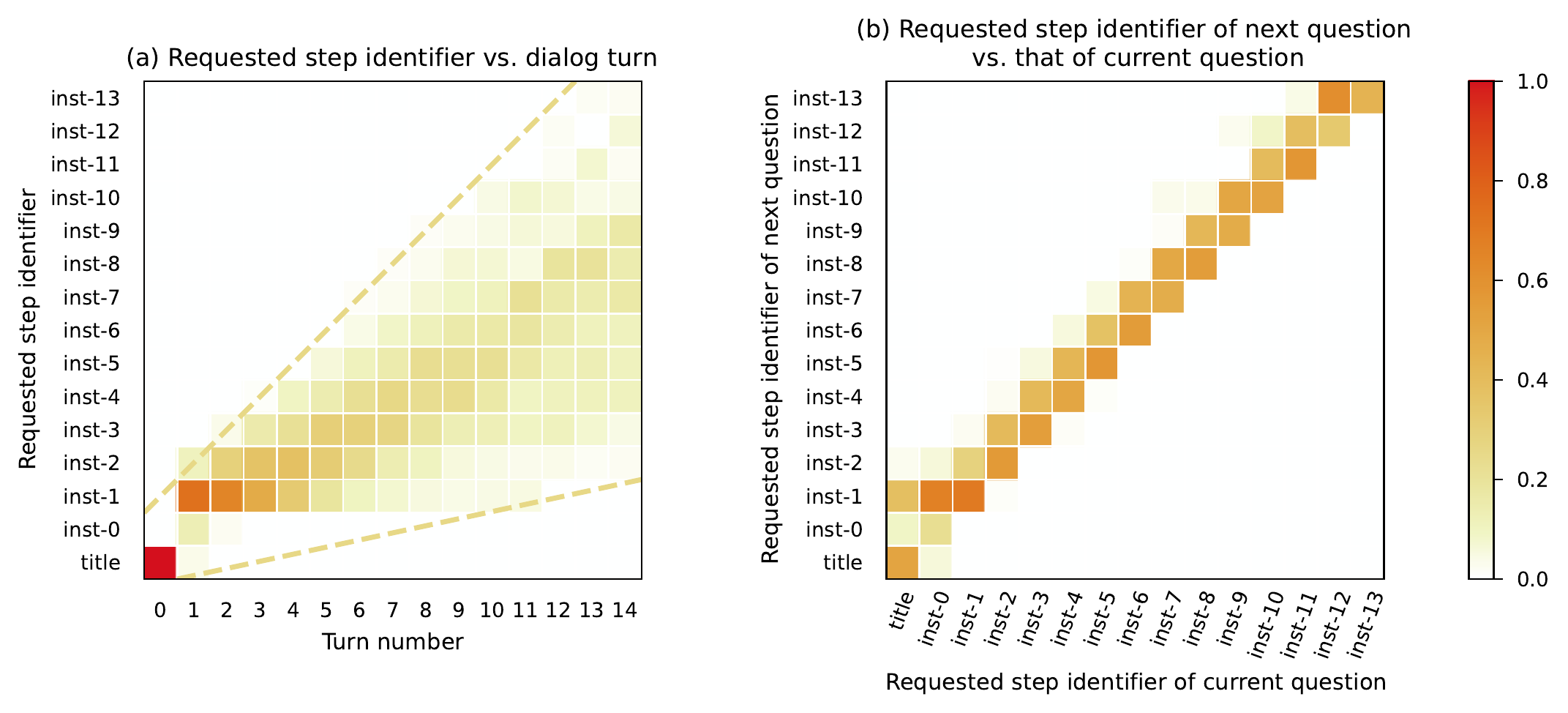}
        \caption[blah]{Heatmaps illustrating the dialogs flow along with the recipe steps.
        \begin{enumerate*}[(a)]
          \item \label{it:heatmap:step-vs-turn} Requested recipe step identifier \vs dialog turn number;
          \item \label{it:heatmap:steps-for-subsequent-turns} The next turn's requested recipe step identifier \vs the current turn's.
        \end{enumerate*}
        (a) and (b) share the same color scale.
        The lengths of dialog turns (\ie a pair of user and agent utterances) and recipe steps are truncated to 15 for simplicity. 
          }
        \label{fig:progress_heatmap}
    \end{figure}

    Since recipes comprise a sequence of ordered steps, the dialog is expected to roughly follow the steps of the corresponding recipe linearly. 
    \figref{fig:progress_heatmap} illustrates such linearity by counting and normalizing the occurrences of 
    \ref{it:heatmap:step-vs-turn} the requested recipe step identifier \vs the dialog turn number, and 
    \ref{it:heatmap:steps-for-subsequent-turns} the requested recipe step identifiers of a pair of two consecutive user questions.
    Note that the recipe step identifiers in \figref{fig:progress_heatmap} refer to a concatenation of 
  the title and the instructions (\ie ingredients are not considered) as the plotted statistics\footnote{We performed vertical normalization on each cell by dividing its frequency by the sum of all the cell frequencies in the same column.}
  were obtained from 86 dialogs in which the agent skips introducing the ingredient list.
    
    Interestingly, most of the mass in the cone-shaped area marked in \figref{fig:progress_heatmap}\ref{it:heatmap:step-vs-turn} lies below the diagonal line.
    If the mass entirely followed the diagonal line, it would imply that the user never has any problem in understanding the agent's utterances and just keeps asking ``What is next?'' through the conversation, which is apparently unrealistic.
    Instead, the user's continuous requests for clarification cause the distribution's mass to diverge from the diagonal line and shift downwards.
    \figref{fig:progress_heatmap}\ref{it:heatmap:steps-for-subsequent-turns} displays the transition between requested recipe step identifiers for the current user question and the next one. 
    The two distinctive diagonal lines reflect that the grounding knowledge for the next question is mostly either in the same recipe step as the current question or in the step immediately following it.
    
  \section{Task Definitions}
  \label{sec:task_definitions}
  
  As stated in \secref{sec:intro}, we focus on solving two challenges during the development of our CA system, 
  \ie challenge~\ref{it:ca-step2}, aligning a conversation and its document, and challenge~\ref{it:ca-step3}, simulating the CA's decision-making process to generate proper responses.
    We introduce three tasks to tackle these challenges as well as to assess the required language understanding abilities in a procedural document setting.
    Note that the challenge~\ref{it:ca-step3} is split into two tasks, \ie \task{2} and \task{3}.
    The definition of each task is listed below.
  
  \subsubsection*{\task{1}: User Question Understanding}
    Understanding a user's question requires the agent to resolve the user's intents and track the dialog state, \ie the annotated \textit{User State frame}.
    Every state frame contains at least one intent $\mathbf{y}_I$ and two state trackers
  pointing to the completed step $\mathbf{y}_C$ and the requested step $\mathbf{y}_R$ respectively. 
    Predicting the requested step can be quite challenging, as it may refer to the dialog context rather than the document.
    For example, the question U2 in \figref{fig:dialog_anno_sample} asks for repeating the first mentioned ingredient in A1.
    
    Formally, the input of \task{1} includes  
    \begin{enumerate*}[(1)]
    \item the user question $Q$,
    \item its dialog history $H$, \ie a given number of utterances preceding the current user question that the CA has to respond to,
  and
    \item the grounding recipe document $D$ including the span boundary indices of step identifiers $\{(s_j^\textit{start}, s_j^\textit{end})\}^{N_S}_{j=1}$ (where $N_S$ denotes the number of step identifiers in a recipe).
    \end{enumerate*}
  (Note that we will experiment with a varying number of preceding turns for the history, see further.)
    The targets are an unordered set of intents ($\mathbf{y}_I$) and two state trackers (completed step $\mathbf{y}_C$, requested step $\mathbf{y}_R$).
    Predicting user intents can be viewed as a multi-label classification problem. 
    The state tracker prediction is formulated as a span selection problem. 
    Unlike the typical question answering task, \eg SQuAD \citep{rajpurkar-etal-2018-know}, our model does not predict the start or end indices of a span from a document.
    Instead, as the recipe step span indices are provided a priori in our setting, we just need to select the most likely one from these given candidate spans.
    Note that for each dialog, the number of span candidates varies with the recipe document.
  
  \subsubsection*{\task{2}: Agent Action Frame Prediction}
  
    As described in \secref{sec:agent_response_annotation}, every agent action frame is composed of 
  agent acts (\eg \formatact{greeting}) and argument pointers, \ie identifiers of the grounding spans.
    Predicting the agent action frame can be viewed as end-to-end modeling of the agent's decision-making process in response to the user's questions.
    Although seemingly similar to \task{1}, \task{2} differs in two aspects: 
   \begin{enumerate*}[(i)]
  \item the sequential order of agent acts (whereas user intent order is less important);
  \item whether or not there need to be argument pointers depends on the agent acts (\ie some acts 
  do not take any arguments).
  \end{enumerate*}
  
    Like \task{1}, the input of \task{2} includes
    \begin{enumerate*}[(1)]
    \item the user question $Q$, 
    \item the dialog history $H$ 
    and \item the recipe document $D$ along with the span indices of the full set of all recipe identifiers $\{(s_j^\textit{start}, s_j^\textit{end})\}^{N_F}_{j=1}$ (where $N_F$ denotes the number of full set of all recipe identifiers in a recipe).
    \end{enumerate*}
    The targets consist of agent acts $\mathbf{y}^\textit{act}$ and the corresponding argument pointers $\mathbf{y}^\textit{frag}$ in which we use a ``dummy span'' 
  to pad those positions without any pointers (also called \emph{null pointers}).
  
  \subsubsection*{\task{3}: Agent Response Generation}
  
    Assuming that the grounding text spans are already given for agent turns,
    this task focuses on generating a natural language response.
    This agent response generation is challenging because
    \begin{enumerate*}[(i)]
    \item in many cases, the model needs to paraphrase the imperative expressions from recipes into different forms 
  (\eg ``Preheat oven to \SI[mode=text]{220}{\degreeCelsius}.''~$\rightarrow$ ``Can you preheat the oven to \SI[mode=text]{220}{\degreeCelsius}?''), and
   \item the system is supposed to resolve the coreferences within a dialog, \eg ``Could you repeat the last ingredient?''. 
   \end{enumerate*}
  
    \task{3} takes as input:
    \begin{enumerate*}[(1)]
      \item the user question $Q$, 
      \item its dialog history $H$, 
      \item a \textbf{Prompt} composed of agent acts of the agent's response, 
    and \item the argument pointer span tokens $G$ as the context.
    \end{enumerate*}
    The target is the agent's gold response.

\section{Baseline Models and Empirical Results}
\label{sec:models_and_results}

  For each task in \secref{sec:task_definitions}, we propose a baseline model leveraging neural networks and pre-trained language models.
  Detailed experimental settings and result analysis can be found in the subsequent subsections.
  For all our baseline models, the empirical results are averaged over 5 runs, each with a different random seed used for model weight initialization and mini-batch sampling. 
  To reduce the impact of selection bias due to the limited number of conversations, a different random partition into train/dev/test sets (distributed as 80\%, 10\%, 10\%) is constructed for each of these runs.
  Since each dialog is linked to a unique recipe text, we ensure that none of the recipes in the dev/test sets are part of the train set. 

  \begin{figure*}
    \centering
    \includegraphics[width=0.95\textwidth]{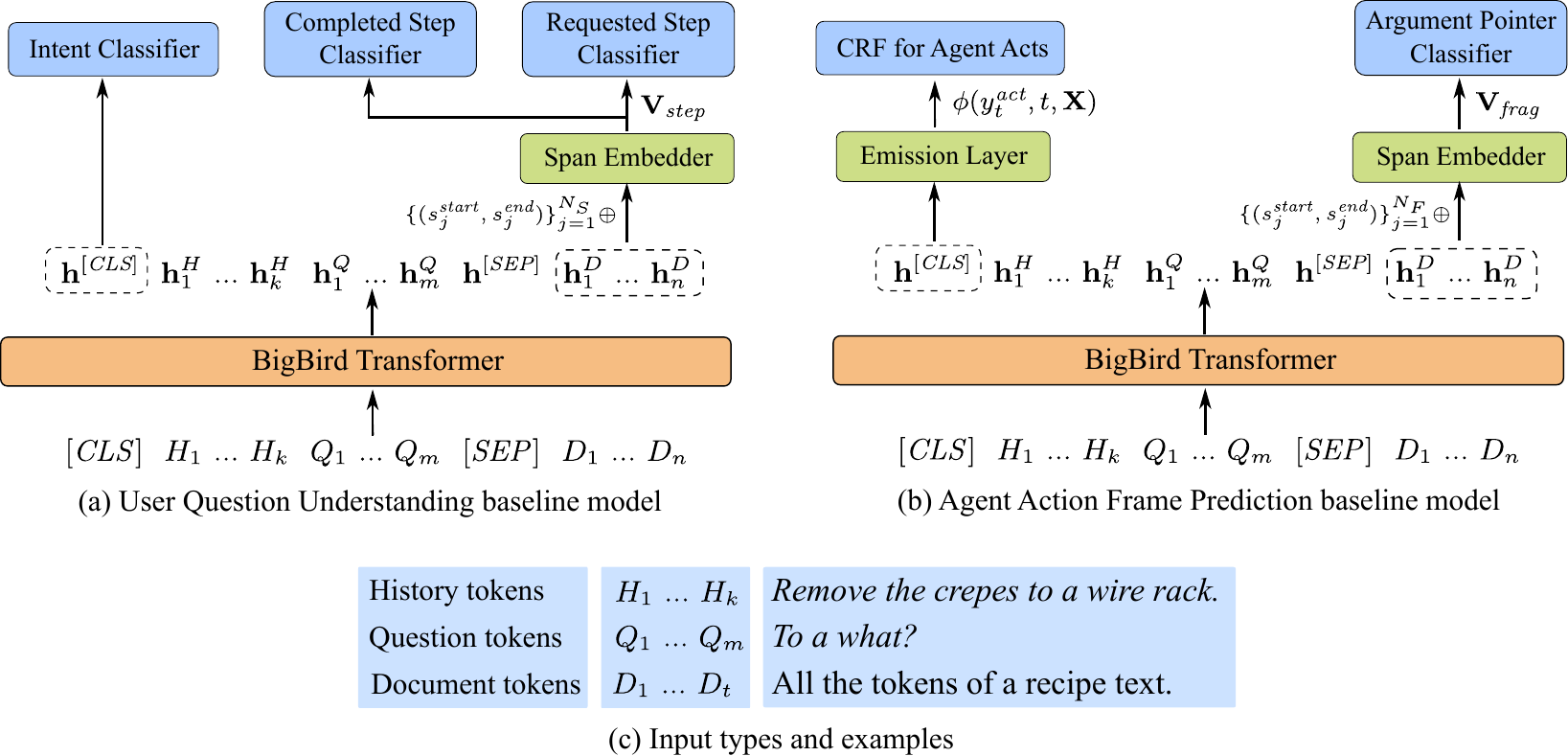}
    \caption{Baseline models for (a)~\task{1}: User Question Understanding (b)~\task{2}: Agent Action Frame Prediction.
            $H, Q, D$ denote tokens of dialog history, user question, and entire recipe text respectively. 
            Example inputs are given in~(c).
            }
    \label{fig:models_user_agent_tasks}
  \end{figure*}

\subsection{User Question Understanding (\task{1})}\label{sec:user_question_understanding}

  \figref{fig:models_user_agent_tasks}(a) sketches our baseline model for \task{1}.
  For all the training instances, the grounding document $D$ is appended after the dialog history $H$ and user question $Q$ in the input. 
  We experiment with the number of history utterances, denoted by \#$H$, ranging from 0 to 10.
  Note that our model in \figref{fig:models_user_agent_tasks}(a) takes the utterances in a chronological order, while doc2dial \citep{feng-etal-2020-doc2dial} also experimented with a reversed setting,
in which the user question and dialog history are concatenated in reversed time order, \ie the latest user question appears first in the input.
  Although doc2dial claimed that the reversed input outperformed the normal one, our pilot experiment results revealed that 
there was no significant difference between these two settings in our dataset.

  The vanilla BERT encoder has a limited input size (\ie there is a 512 token limit), which does not suffice to contain longer full documents. 
  A common solution to circumvent this limitation is to segment the long input document and adopt a sliding window approach to do so~\citep{feng-etal-2020-doc2dial, Qu2019AttentiveHS}.
  Since this means that a single original training instance is converted to multiple ones each covering part of the long input document, this implies a non-negligible computational overhead.
  Simply truncating the document would clearly lead to discarding potentially useful input information.
  Therefore, we use BigBird transformer~\citep{zaheer2020big} as the encoder, which has the advantage of sparse attention, which reduces the computation complexity from $O(n^2)$ to $O(n)$ in case of long inputs.
  As a result, the model can handle long documents more efficiently.
  Training instances are fed in batch to the encoder, which transforms tokens to hidden representations $\mathbf{h} \in \mathbb{R}^{b \times m \times s}$, as per \equref{eq:encoder_transform}; here,
  $b, m, s$ denote the batch size, length of input tokens and hidden size.
  We use the [CLS] token embedding $\mathbf{h}^{[\textit{CLS}]} \in \mathbb{R}^{b \times m \times s}$ to represent the query vector.
  The document token embeddings $\{\mathbf{h}^{D}_{i}\}^{N_D}_{i=1}$ are passed to a span embedder ($N_D$ is the total number of document tokens), which concatenates the start and end token embeddings indexed by 
the step span tuples $\{(s_j^\textit{start}, s_j^\textit{end})\}^{N_S}_{j=1}$ to form step span vectors $\mathbf{V}_\textit{step} \in \mathbb{R}^{N_S \times 2s}$. 
  \begin{align}
    \mathbf{h} &= \text{BigBird}(H, Q, D) \label{eq:encoder_transform} \\
    \mathbf{V}_{step} &= \text{SpanEmbedder}(\{\mathbf{h}^{D}_{i}\}^{N_D}_{i=1}, \; \{(s_j^{start}, s_j^{end})\}^{N_S}_{j=1})
  \end{align}
  On top of the encoder, a feed-forward neural network (FFNN)\footnote{By default, all FFNNs in this work are composed of 1~hidden layer activated by the GELU function and 1 output layer.} maps $\mathbf{h}^{[\textit{CLS}]}$ to a logit vector of which each dimension represents the unnormalized classification score of an intent. 
Given the multi-label nature of the user intent classification task, the probability $\hat{p}_{I}$ of each intent $I$ is modeled by applying a sigmoid function on the corresponding logit, denoted by \equref{eq:intent_prob}.
  After the span embedder, span vectors are sent to two independent FFNNs and softmax layers that predict the probability distributions $\hat{p}_c$, $\hat{p}_r$
for completed and requested state trackers, respectively, shown in \equsref{eq:completed_step_prob}{eq:requested_step_prob}.
  \begin{align}
    \hat{p}(I \vert X; \theta) &= \sigmoid(\text{FFNN}(\mathbf{h}^{[\textit{CLS}]})) \label{eq:intent_prob} \\
    \hat{p}(C \vert X; \theta) &= \softmax(\text{FFNN}(\mathbf{V}_\textit{step})) \label{eq:completed_step_prob} \\%
    \hat{p}(R \vert X; \theta) &= \softmax(\text{FFNN}(\mathbf{V}_\textit{step})) \label{eq:requested_step_prob}
  \end{align}
  During training, the model computes the cross entropy losses of the 3 probability distributions against 
their corresponding target distributions $p(\mathbf{y}_I), p(\mathbf{y}_C), p(\mathbf{y}_R)$.
  We implement joint training by minimizing the sum of three losses.
  For a given training instance $\mathbf{X}$, the local loss is:
  \begin{equation}
      \mathcal{L} = - \left[p(\mathbf{y}_I) \cdot \log \hat{p}(I \vert X; \theta) 
                  + p(\mathbf{y}_C) \cdot \log \hat{p}(C \vert X; \theta)
                  + p(\mathbf{y}_R) \cdot \log \hat{p}(R \vert X; \theta)\right]
  \end{equation}
  We use the F1 score for the intent prediction evaluation. 
  The performance of state tracking is evaluated using the accuracy metric.

\subsubsection*{Experimental Results}
\begin{figure*}[ht]
    \centering
    \includegraphics[width=0.95\textwidth]{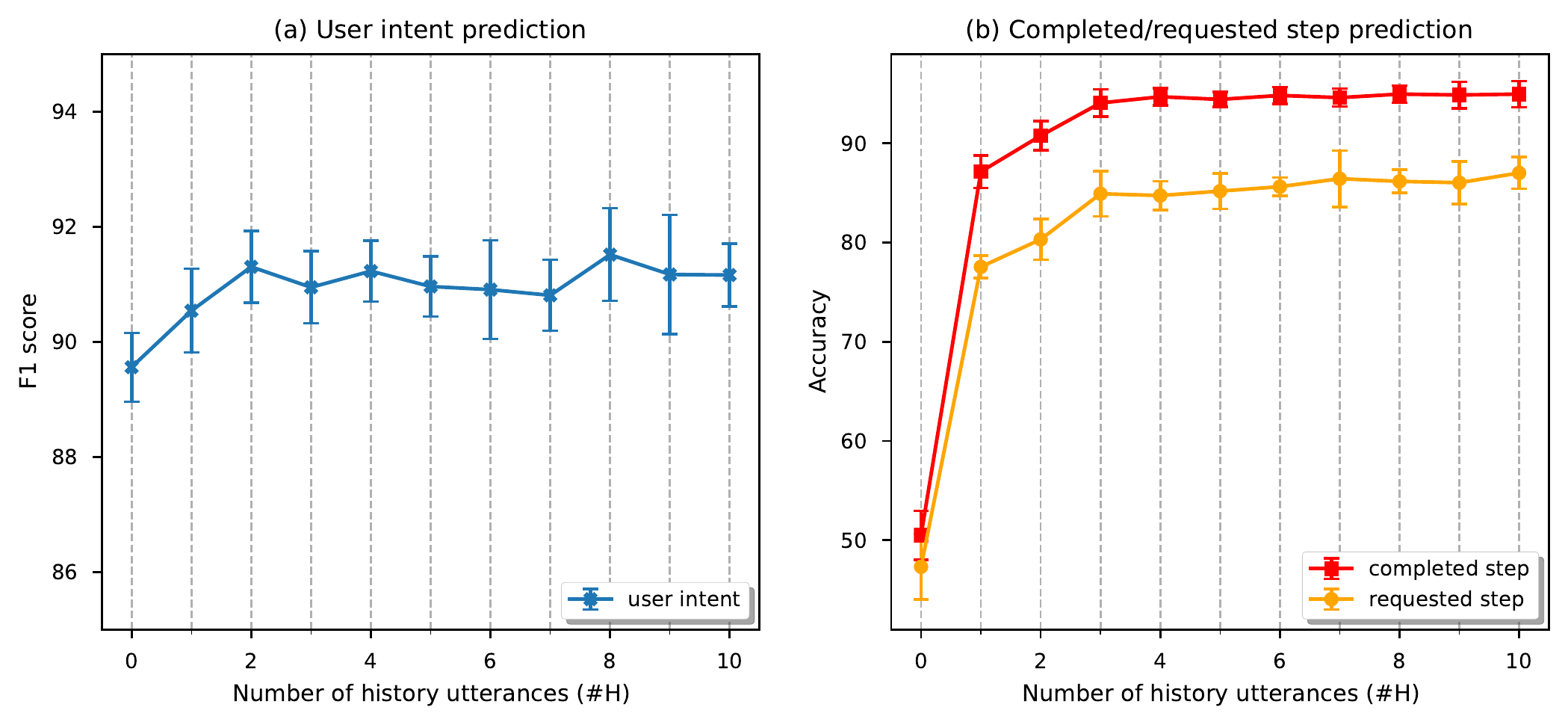}
    \caption{Experiment results of \task{1} User Question Understanding.
            The results are averaged over 5 runs with different randomly sampled splits of CookDial.}
    \label{fig:user_task_results}
  \end{figure*}

  From \figref{fig:user_task_results}, we note that
  the number of history utterances has less influence on the intent prediction than that on the accuracy of state tracking.
  The F1 score of predicting user intents stays around 91\% for the test set. 
  This stationary performance suggests that a single user question is informative enough for the model to infer the user's intent.
  The fact that there is no drastic drop of the F1 score when the history length is 0 also validates this assumption.

  As for state tracking on the test set, the accuracy increases to 86.4\% and 94.5\% for the requested step and completed step tracker, respectively, 
by adding more history utterances. 
  However, the F1 scores of both step tracker predictions start plateauing from \#$H$=3 onward.
  This implies that, as intuitively expected, understanding a user question benefits little from its distant context.
  We note that predicting the completed step reaches higher performance than the requested step, which stems from the fact that the user's request can be rather ambiguous. 
  For example, when the user is asking ``What is next?'', they may request an ingredient or instruction depending on the dialog context.
  Further, as expected, without any history (\#$H$=0) the model largely fails at state tracking (with accuracy dropping to around 50\%).


\subsection{Agent Action Frame Prediction (\task{2})} \label{sec:agent_action_frame_prediction_task}

  \figref{fig:models_user_agent_tasks}(b) gives an overview of the baseline model for \task{2}.
  We again use BigBird as the encoder but fine-tune it independently from \task{1}.
  The span embedder creates a zero vector $\mathbf{V}_z$ as the dummy span embedding to represent null pointers. 
  In addition, given the argument span tuples, the final output of the span embedder is $\mathbf{V}_\textit{frag}=\{\mathbf{V}_z, \hat{\mathbf{V}}_\textit{frag}\}$, where $\hat{\mathbf{V}}_\textit{frag}$
is an end-point span vector. 
  \begin{align}
    \mathbf{h} &= \text{BigBird}(H, Q, D) \label{eq:encoder_agent_act} \\
    \mathbf{V}_\textit{frag} &= \{\mathbf{V}_z, \hat{\mathbf{V}}_\textit{frag}\}, \nonumber \\
    &\hspace{-1em}\text{ where} \; \hat{\mathbf{V}}_\textit{frag} = \text{SpanEmbedder}\left(\{\mathbf{h}^{D}_{i}\}^{N_D}_{i=1}, \; \{(s_j^\textit{start}, s_j^\textit{end})\}^{N_F}_{j=1}\right)
  \end{align}
  Again, the [CLS] token embedding $\mathbf{h}^{[\textit{CLS}]}$ is used as the query vector.
  To predict the agent act sequence, we use conditional random fields (CRFs)~\citep{Sutton2012AnIT} to model the dependency between agent acts. 

  Since in our dataset we observe at most 4 agent acts in the agent response, we simplify the model by fixing the CRF output length as 5 (appending an <eos> token after an act sequence). 
  The potential function in the CRF layer is decomposed into an emission function $\phi$ and a transition function $\psi$. 
  For the emission function $\phi$, we train an FFNN layer to transform the contextual vector $\mathbf{h}^{[\textit{CLS}]}$ into the emission scores $\phi(y^\textit{act}_k, k, \mathbf{X})$.
  The transition function $\psi$ is represented by a trainable matrix $\mathbf{W}_{s}$ maintaining the transition score between agent act tags and two extra tags \ie <start> and <end>.
  Finally, the predicted probability for an agent act sequence is computed as \equref{eq:agent_act_prob}.
  As for the argument pointer prediction, we pass the argument pointer span vector $\mathbf{V}_\textit{\,frag}$ to another FFNN layer. 
  The output logit is then activated by a softmax function to predict the probability distribution $p(\mathbf{y}^\textit{frag}\vert\mathbf{X};\theta)$ of the argument pointers, as shown in \equref{eq:argument_pointer_prob}.
  \begin{align}
    \phi(y^\textit{\,act}_k, k, \mathbf{X}) &= \text{FFNN}(\mathbf{h}^{[\textit{CLS}]}), 
    \quad \psi(y^\textit{\,act}_{k-1}, y^\textit{\,act}_k) = \left(y^\textit{\,act}_{k-1}\right)^T \, \mathbf{W}_{s} \, y^\textit{\,act}_{k}\label{eq:emission_for_crf}\\
    \hat{p}(\mathbf{y}^\textit{\,act}\vert\mathbf{X};\theta) &= \frac{1}{Z(\mathbf{X})} \exp\left(\sum_{k=0}^\tau\phi(y^\textit{\,act}_k, k, \mathbf{X}) + 
                                                                \sum_{k=1}^{\tau-1}\psi(y^\textit{\,act}_{k-1}, y^\textit{\,act}_k)\right)\,\mathbf{,}  \notag\\
        &\hspace{-3em}
        \text{where} \; Z(\mathbf{X})= \sum_{\tilde{y}^\textit{\,act}}^{\mathcal{Y}^\textit{\,act}} \exp\left(\sum_{k=0}^{\tau}\phi(\tilde{y}^\textit{\,act}_k, k, \mathbf{X}) + 
                                                            \sum_{k=1}^{\tau-1}\psi(\tilde{y}^\textit{\,act}_{k-1}, \tilde{y}^\textit{\,act}_k)\right) \label{eq:agent_act_prob} \\
    \hat{p}(\mathbf{y}^\textit{frag}\vert\mathbf{X};\theta) &= \softmax(\text{FFNN}(\mathbf{V}_\textit{frag})) \label{eq:argument_pointer_prob}
  \end{align}
  For one training instance, the local loss is computed by summing the negative log likelihood of $\hat{p}(\mathbf{y}^{\,act}\vert\mathbf{X};\theta)$ 
and the cross entropy of the predicted pointer likelihood against its target distribution $p(\mathbf{y}^\textit{\,frag})$:
  \begin{equation}
        \mathcal{L} = -\left[\,\log \hat{p}(\mathbf{y}^\textit{\,act}\vert\mathbf{X};\theta) + p(\mathbf{y}^\textit{frag}) \cdot \log \hat{p}(\mathbf{y}^\textit{\,frag}\vert\mathbf{X};\theta)\,\right]
  \end{equation}
  For this task, we investigate the influence of the number of history utterances.
  Both of the act sequence and argument pointer predictions are evaluated by F1 scores.

\subsubsection*{Experimental Results}

\begin{figure*}[ht]
    \centering
    \includegraphics[width=0.6\textwidth]{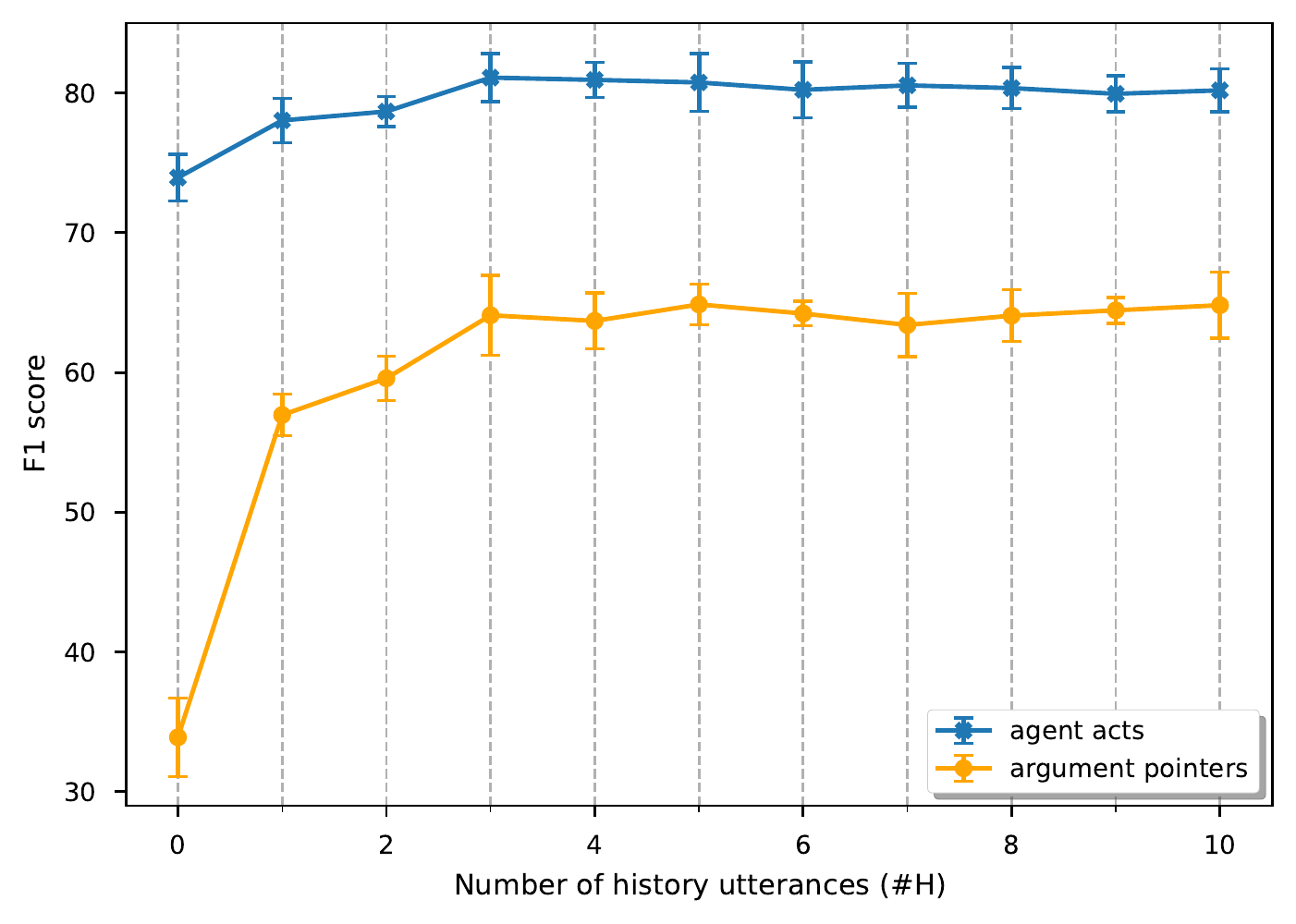}
    \caption{Experiment results of \task{2} Agent Action Frame Prediction. 
    All the numbers are calculated over 5 random seeds.}
    \label{fig:agent_task_results}
  \end{figure*}

  Prediction results in \figref{fig:agent_task_results} show that the length of dialog history more significantly impacts the performance of argument pointer prediction (F1 score drops by \textgreater\,30\% for \#$H$ decreasing from 5 to 0) than agent act prediction (where F1 score drops by $\sim$\,6\%).
  This difference is expected, since predicting the agent acts mainly relies on good understanding of the user intent embedded in the previous question
and is less history dependent.
  For example, if the user intents are ``\formatintent{confirm, req\_instruction}'', we can presume an act sequence like ``\formatact{inform\_instruction}'' as the most likely prediction
without knowing extra information from distant utterances. 
  In contrast, predicting the argument pointer sequence is strongly dependent on at least some history, since the model needs to update the user state (\ie cooking progress)
when the user is asking about a distant entity. 
  When \#$H$ exceeds 3, the model performance plateaus, suggesting that it is not necessary to include all of the dialog history as input in our system.
  
\subsection{Agent Response Generation (\task{3})} \label{sec:agent_response_generation_task}

\begin{figure*}[t!]
  \centering
  \includegraphics[width=0.8\textwidth]{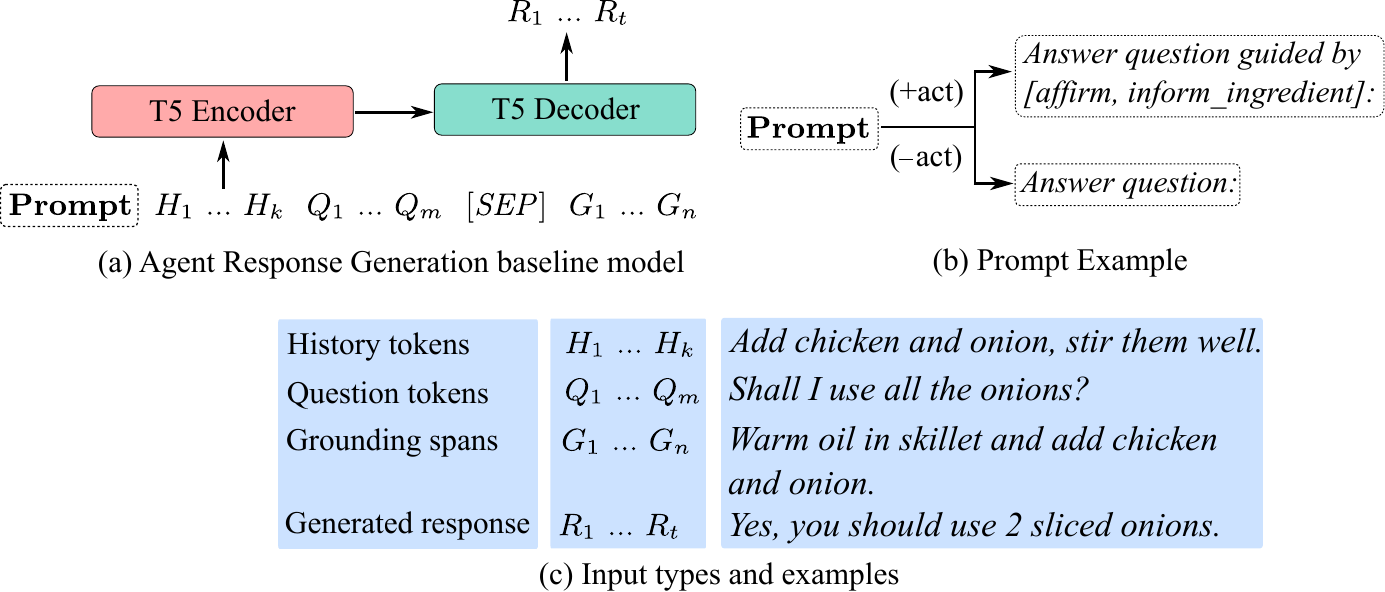}
  \caption{(a)~Baseline model of \task{3}: Agent Response Generation.
  (b)~Examples of dynamic and static \textbf{Prompt} depending on whether agent acts are used or not. 
  (c)~Input and output examples.}
  \label{fig:models_generation_task}
\end{figure*}

  For the answer text generation, we rely on the pretrained T5-base model \citep{T5model-Raffel2020ExploringTL}, which adopts a unified ``text-to-text'' approach casting NLP tasks as text generation and currently attaining state-of-the-art results on many of them. 
  The pretrained encoder and decoder of T5-base are finetuned during our training.
  We use beam search with a beam width of 10 during  decoding.
  \figref{fig:models_generation_task}(a) illustrates our baseline model.
  For the dialog history, we prepend the history tokens to the current question.
  The number of history utterances \#$H$ is fixed at 5 for all the models.
  Our model is designed to tackle the two challenges mentioned in \secref{sec:agent_response_generation_task} (\ie paraphrasing and coreference resolving) by leveraging the semantic information from \emph{Agent Action Frames}.
  Firstly, we hypothesize that generating a proper response can be enhanced
by introducing the agent's dialog acts into the model input. 
  More specifically, we integrate the agent acts into the prompt of the T5 model to inject essential semantic information,
  which is denoted as the (\textsf{$+$act}) setting as shown in \figref{fig:models_generation_task}(b).
  We also investigate another setting in which prompts contain no agent dialog act information, named as (\textsf{$-$act}) in our experiments.
  In the latter case (\textsf{$-$act}), the prompt remains static for all the training instances, \ie ``Answer question''.
  In contrast, the (\textsf{$+$act}) setting makes the prompt dynamic, since the agent acts vary among agent utterances in a conversation.
  For example, the dynamic prompt used in \figref{fig:models_generation_task}(b) is ``Answer question guided by [affirm, inform\_ingredient]''.
  Secondly, identifying the needed information from a long recipe is actually highly dependent on the grounding part, \ie $G$ in \figref{fig:models_generation_task}(a).
  For the argument pointer spans, we also experiment with two settings: 
using gold argument pointer spans (\textsf{$+$pointer}) or not (\textsf{$-$pointer}).
  For the former, $G$ is in fact composed of the recipe text spans denoted by argument pointers within an \emph{Agent Action Frame}.
  In the (\textsf{$-$pointer}) setting, we replace the gold spans with the entire recipe text, which makes it hard for the model to locate the key information precisely.
  Since the combination of  (\textsf{$+$act}, \textsf{$+$pointer})  brings minimal noise into the input, we also call it the oracle model.
  We perform ablation experiments to asses the influence of providing agent dialog acts and argument pointers as input to the generation model.
  Given a training instance $\mathbf{X}$ and its paired response $\mathbf{Z}$, the model minimizes cross-entropy loss:
  \begin{equation}
      \mathcal{L} = -\sum_{t=1}^{\mathbf{Z}}\log p(z_t \vert z_{t-1}, \ldots,z_1,\mathbf{X};\theta)
  \end{equation}
    
\subsubsection*{Experimental Results}
  We adopt BLEU-1/2/3/4 and ROUGE-L as automatic evaluation metrics for the answer generation task.
  \Tabref{tab:generation_task_results} reports the performance of our oracle model and its ablated variants on test set.
  Not surprisingly, the oracle model is the most competitive with 36.5 BLEU-4 and 54.4 ROUGE-L. 
  Ablation of agent acts and argument pointer spans individually (row 2 and 3) shows little impact on the BLEU-4 score, while BLEU-1 and ROUGE-L results are more clearly affected.
  The larger decrease of ROUGE-L when agent acts are not used (\textsf{$-$act}, \textsf{$+$pointer}) implies that excluding the agent acts from the model input could harm
the model's ability to generate responses or copy phrases from the document in a more appropriate way (see generation samples in \secref{sec:generation_samples}).
  Moreover, when we remove both agent acts and pointer spans (\textsf{$-$act}, \textsf{$-$pointer}), the sharp drop of performance suggests 
the necessity of agent acts and document spans as inputs to improve the response generation quality of a document-grounded dialog system.

  \begin{table}[htbp] \small \centering
    \caption{Experiment results of \task{3}~Agent Response Generation. 
        The results are averaged over 5 runs with different randomly sampled splits of CookDial.
        ``act'' and ``pointer'' denote agent acts and argument pointer spans respectively.} 
    \label{tab:generation_task_results}
    \begin{tabular}{lccc}
      \toprule
                & Settings             & BLEU-1/4                          & ROUGE-L    \\
      \midrule
      1 &(\textsf{$+$act}, \textsf{$+$pointer}) oracle & \textbf{54.2$\pm$0.6 / 36.9$\pm$0.6}    & \textbf{54.2$\pm$0.5}\\ 
      2 &(\textsf{$-$act}, \textsf{$+$pointer})     & 51.6$\pm$0.8 / 35.3$\pm$0.6   & 51.0$\pm$0.6        \\ 
      3 & (\textsf{$+$act}, \textsf{$-$pointer})     & 52.6$\pm$1.8 / 36.08$\pm$1.4   & 52.54$\pm$1.6         \\ 
      4 &(\textsf{$-$act}, \textsf{$-$pointer})     & 46.2$\pm$0.4 / 31.10$\pm$0.5   & 46.31$\pm$0.9 \\ 
      \bottomrule
    \end{tabular}
  \end{table}

\subsubsection*{Case study on generated response samples}\label{sec:generation_samples}

  \Tabref{tab:case_study_turn_samples} compares the generated responses sampled from two models, \ie the (\textsf{$-$act}, \textsf{$-$pointer}) model and oracle model (\textsf{$+$act}, \textsf{$+$pointer}). 
  In general, the oracle model generates more coherent responses than the ablated model does, which complies with our hypothesis that 
introducing the agent dialog acts into the model input can enhance the generation quality.
  On the other hand, the agent act is found to have great impact on steering the generative model.
  For example, in questions~1 and 4, the act ``\formatact{affirm}'' helps the oracle model to produce affirmative phrases
like ``yes, great'' to respond smoothly to the user's yes/no questions, \eg ``Is all-purpose flour good?''.
  In questions~2 and 3, without the guidance of agent acts, the ablated model gives totally wrong answers,
which may confuse the user or even lead to a failure task in real applications.
  However, even when integrating agent acts (\textsf{$+$act}), the task remains highly challenging and responses are not perfect.
  In question~1, the oracle model ignores the act ``simplify\_continue'' and does not mention the next ingredient ``white sugar''.
  In question~5, although the oracle model understands the meaning of ``\formatact{sum\_ingredient\_list}'', the total number is wrongly calculated.
  This numerical mistake reveals that our model lacks the ability to reason over the numerical information in the recipes. 
  Mitigating this is left for future work.
  
  The dialog snippets in \Tabref{tab:case_study_dialog_samples} are consecutive turns generated by the two same models in \Tabref{tab:case_study_turn_samples}.
  Without the assistance of \textit{argument pointers}, the ablated model's generated responses are more erroneous and inconsistent.
  Compared to the ablated model, the oracle model succeeds in finding correct references for both user questions, 
\ie ``the beginning'' and ``what kind of sugar''.
  This empirical observation implies that our \textit{argument pointer} annotation helps the model to 
alleviate the difficulty in finding coreferences within dialogs. 
  However, both models fail to answer the last question correctly, which requires the model to resolve long-distance anaphora.
  This limitation will be further discussed in \secref{sec:conclusion}.
  \newpage
  \begin{landscape}
  \begin{table}[htbp] \small
    \caption{Generated responses sampled from different experimental settings. 
    The blue and italic font indicates a wrong response.} 
    \label{tab:case_study_turn_samples}
    \begin{tabular}{p{3cm} p{4cm} p{2.4cm} p{3.5cm} p{3.5cm}} 
      \toprule
        Question     & Gold Response & Agent Acts & Responses generated by (\textsf{$-$act}, \textsf{$-$pointer}) model  & Responses generated by (\textsf{$+$act}, \textsf{$+$pointer}) model   \\
      \midrule
      1. What kind of flour should it be? Is all-purpose flour good? & Yes it is! Also add 1 cup of white sugar. & \vtop{\hbox{\strut \formatact{affirm();}} \hbox{\strut \formatact{simplify\_continue(ing-1);}}} &  \textitblue{Sorry I cannot answer that question.} & Yes, it should be all-purpose flour.  \\ 
      \hline
      2. Should I add all of the egg? &  I cannot answer that question.  & \formatact{fence();}  & \textitblue{Yes, you should add all of the eggs.} & Sorry I cannot answer that question. \\ 
      \hline
      3. Something else to do still before putting them in the oven? & No that is it! & \formatact{negate();} & \textitblue{You can now bake it for 10 to 12 minutes.} & No. \\
      \hline
      4. Ok no problem, can we start with the recipe? & Yes let us start! Do you have a slow cooker? & \vtop{\hbox{\strut \formatact{affirm();}} \hbox{\strut \formatact{verify(tool-0-0);}}} & First place the chicken in a slow cooker. & Great! Can you take a slow cooker?  \\
      \hline
      5. Okay, what is the first thing I have to do? & There are 12 ingredients. Do you need me to list the ingredients? & \vtop{\hbox{\strut \formatact{sum\_ingredient\_list();}} \hbox{\strut \formatact{propose\_list\_ingredients();}}} & \textitblue{Preheat the oven to 175 degrees Celsius.} & \textitblue{This recipe has 9 ingredients.} Do you want to prepare them now? \\
      \bottomrule
    \end{tabular}
  \end{table}

  \begin{table}[t] \small 
    \caption{Dialog snippets from different experimental settings.
    The blue and italic font indicates a wrong response.} 
    \label{tab:case_study_dialog_samples}
    \begin{tabular}{p{0.8cm} >{\raggedright}p{5cm} >{\raggedright}p{5cm} >{\raggedright}p{5cm}} 
      \toprule 
      Role  & Gold dialog  & Dialog generated by (\textsf{$-$act}, \textsf{$-$pointer}) model & Dialog turn generated by (\textsf{$+$act}, \textsf{$+$pointer}) model    \\
      \midrule
      \textit{Agent} & Good. The next two are one and one third cup of white sugar and 2 eggs. & The next two ingredients are 1 1/3 cup of white sugar and 2 eggs. & The next two ingredients are 1 1/3 cup of white sugar and 2 eggs. \\ 
      \hline
      \textit{User}  & \multicolumn{3}{c}{Could you repeat the beginning please?} \\ 
      \hline
      \textit{Agent} & The first ingredient is 1 and 1/3 cup of white sugar. & \textitblue{2 eggs.} & 1 and 1/3 of a cup of white sugar. \\
      \hline
      \textit{User}  & \multicolumn{3}{c}{Ok. Thanks for the clarification. What kind of sugar should it be?} \\
      \hline
      \textit{Agent} & It should be white sugar.  & \textitblue{sorry I cannot reply to that question.} & White sugar. \\
      \hline
      \textit{User}  & \multicolumn{3}{c}{You said 2 eggs after that, right?} \\ 
      \hline
      Agent & Yes, you also need to add 2 eggs & \textitblue{1 tablespoon of baking powder.} & Yes, \textitblue{you need 1 1/3 cup of white sugar.} \\
      \bottomrule
    \end{tabular}
  \end{table}
\end{landscape}
  \section{Conclusion and Future Work}
  \label{sec:conclusion}
    The main objective of this paper was to present a new dataset to facilitate research on document-grounded conversational agents (CAs) to assist in procedural tasks.
    Existing works on document-grounded CAs lacked the procedural aspect of the supporting documents, \ie the fact that they are organized in a sequence of steps wherein entities undergo state changes 
  that the CA needs to be aware of and be able to relate to one another. 
    Indeed, for a CA it is crucial to relate the dialog context (\eg which procedural step the user has arrived at) to the prescribed procedure
  to be able to generate an appropriate utterance in the conversation.
    Our newly created CookDial dataset contains 260 human-to-human dialogs that were collected based on recipe documents.
    We proposed an annotation approach for dialog systems grounded in procedural knowledge, with a schema that enables entity and state alignment between dialogs and corresponding documents. 
    Our resulting annotated conversations exhibit non-trivial agent decision-making behavior, including responses containing a varying number of agent acts, segmentation of long instructions and paraphrasing the source document texts.
    From the CA perspective, we identified three major tasks, for which we also established baseline solutions: 
  \begin{enumerate*}[(i)]
      \item user question understanding,
      \item agent action frame prediction, and
      \item agent response generation.
  \end{enumerate*}
    We publicly release the dataset and the baseline models to spur further research.
    In terms of next steps, we highlight three directions that will guide our future work.

    First, we want to explore generalization of our annotation method by applying it to other types of procedural tasks.
    Two more technical application domains of interest include chemical and mechanical operations.
    In these different domains, most definitions in our dialog-act taxonomy (\Tabref{tab:dialog_acts}) will still be useful like ``\formatintent{req\_instruction, req\_subsititute}''.
    However, chemical processes are extremely sensitive to the element quantities, which raises the importance of accurate numerical understanding for the dialog system.
    On the other hand, in mechanical manuals, the spacial relation plays a significant role, \eg in complex assembly tasks. 
    Such mechanics related conversations may need intent annotations about the relative position between objects (\eg ``\formatintent{req\_distance}'').
    To facilitate such more refined acts, we envision a hierarchical structure of dialog acts to develop a cross-domain system. 
    The hierarchy would start with domain-agnostic acts, then branch into domain-specific acts.
    Furthermore, it will be of great value to train a general language understanding model that can be applied for different procedural domains.
  
\begin{figure}
      \centering
      \caption{A dialog snippet illustrating ambiguous anaphora across utterances.}
      \label{fig:hard_example}
      \includegraphics[width=0.8\textwidth]{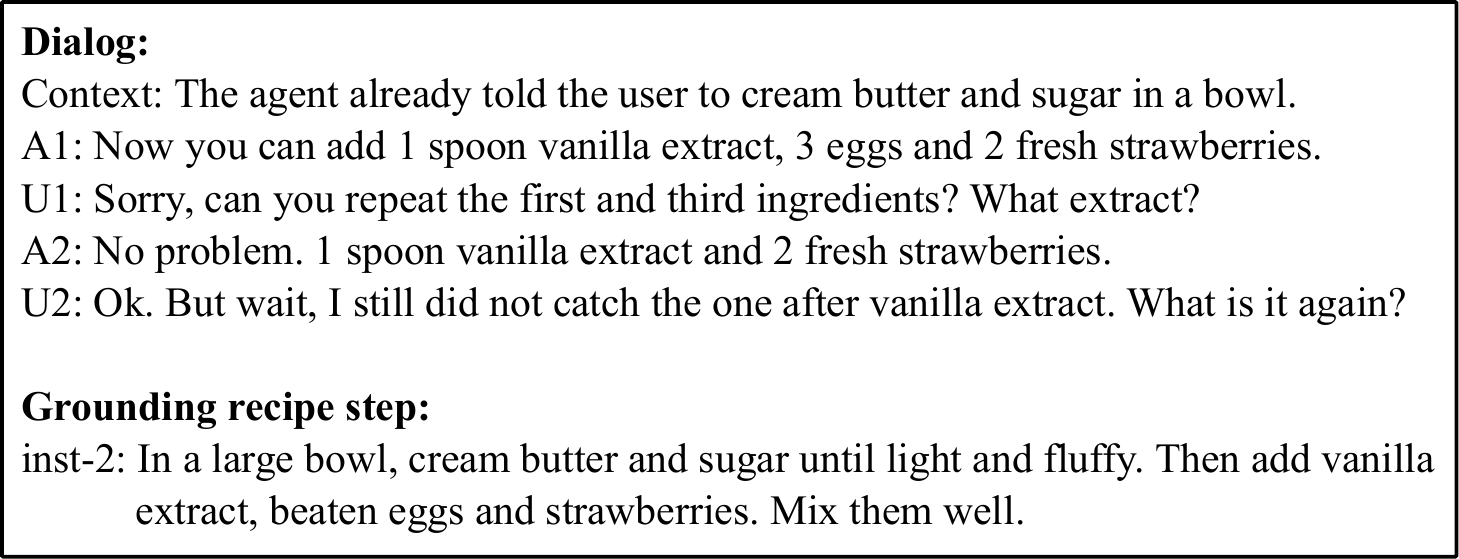}
    \end{figure}

    Second, our current solution lacks the ability to resolve ambiguous anaphora, which makes some user questions particularly hard to answer.
    A dialog snippet in \figref{fig:hard_example} illustrates this.
%
    For question U1, the system needs to understand that the order reference ``first and third'' relates to the preceding utterance A1, rather than the ingredient order in the grounding recipe step.
    The user would be very confused if the agent answers wrongly with the first and last ingredients from \formatid{inst-2}, \ie ``butter and vanilla extract''.
    Similarly, the subsequent user question U2 with ``the one after vanilla extract'' is even more ambiguous, and the CA should attach the context of A2 (rather than A1) to resolve it.
    Our current annotation relies on the absolute identifiers from the recipe,  
  while coreferences within the dialog are not annotated (\eg linking ``the one after vanilla extract'' to the preceding agent utterance's ``2 fresh strawberries'').
    To overcome this problem, we envision extra annotations of coreference linking across utterances.

    Third, we note that in our collected dialogs, users sometimes tend to ask for clarifications on items (ingredients, tools) that they are not familiar with.
    For now, our system performs poorly on answering those questions, given that the dialog agent's knowledge (besides the conversation itself) is limited to the grounded document. 
    The most common answer observed in our collected dataset is ``Sorry, I cannot answer your question''.
    These uninformative answers normally do not disrupt the dialog flow but might 
    damage the user satisfaction with (and therefor limit adoption of) the CA solution.
    This problem can be overcome by incorporating external knowledge sources (\eg Wikipedia) or knowledge graphs (\eg DBpedia).
    How to efficiently fuse such knowledge base information with grounded documents is another direction for our future work.

  
  \section*{Acknowledgements}
    We thank  Maarten De Raedt and Amir Hadifar for their insightful suggestions in the initial data collection.
    The first author is supported by \textit{China Scholarship Council} (No.\ 201906020194) and \textit{Bijzonder Onderzoeksfonds (BOF) van Universiteit Gent} (No. 01SC0618). 
    This research also receives funding from the Flemish Government under the \textit{``Onderzoeksprogramma Artifici\"{e}le Intelligentie (AI) Vlaanderen''} programme.


\begin{appendices}
  
  \section{Experiment Settings} \label{sec:appendix_experiment_settings}
    All the transformer modules in our models are implemented with the Huggingface library \citep{wolf-etal-2020-transformers}.
    We conducted the experiments with a single Nvidia-Tesla-V100 (32GB) card.
    For all the tasks, we use the AdamW optimizer \citep{loshchilov2018decoupled}. 
  For both of \task{1} and \task{2}, we use two different learning rates depending on the layers to accelerate convergence:
  \begin{enumerate*}[(i)]
      \item $10^{-5}$ for the layers within the BigBird encoder;
      \item $10^{-3}$ for the top classifier layers (FFNNs and CRF).
  \end{enumerate*}
  For \task{3}, the learning rate for all the layers is set to ${3\times10^{-4}}$. 
    The batch size is set to 8. 
    The hidden size for all the FFNN layers is 128 except the intent classifier layer (64) in \task{1}.
    The dropout is set to 0.2 in the fine-tuning when needed.

  \section{User Intent and Agent Act Annotations} \label{sec:appendix_annotation_details}
  \setcounter{table}{0}
  \renewcommand{\thetable}{\Alph{section}.\arabic{table}}
    Elucidation on how we annotate the user intents and agent acts is presented in \Tabref{tab:detailed_user_intent} and \Tabref{tab:detailed_agent_acts} respectively.
    For each intent or agent act, we also provide an annotation example except a few, \ie \formatintent{other}, \formatintent{repeat}. 

    \begin{longtable}{p{3cm} p{4cm} p{4cm}}
        \caption{Annotation scheme for the user intents.} \label{tab:detailed_user_intent} \\
        \toprule
          User Intent     & Meaning       & Examples \\
        \midrule
      \endfirsthead

        \caption{(continued) Annotation scheme for the user intents.} \\
        \toprule
          User Intent     & Meaning       & Examples \\
        \midrule
      \endhead

        \midrule
          \multicolumn{3}{r}{\footnotesize\itshape Continue on the next page}
      \endfoot

        \bottomrule
      \endlastfoot
          greeting        & User says ``Hi'' or ``Hello'' to start a conversation.    & Hi! / Hello! / Good morning! \\
          thank           & User expresses gratitude.     & Thank you. / Thanks. \\
          confirm         & User establishes the fact that he or she has accomplished one or several actions. & Done. / I have made it. / All of them are taken out of oven. \\
          negate          & User gives a negative statement.  & No, I don't have any avocado. / Sorry, I cannot remove the cover. \\
          affirm          & User gives a positive statement.  & Yes, I have an electrical beater. / Of course, lasagne is my favorite. \\
          goodbye         & User expresses good wishes at the end of a conversation.  & Bye. /  Goodbye. / See you. \\
          req\_start      & User requests the agent to give the first guidance.   &  How should I start? / How do I make the secret pie? \\
          req\_temperature & User requests the exact temperature expression for an action.  & What temperature shall I set for the baking? \\ 
          req\_instruction & User requests the next instruction after he or she confirms accomplishment of previous instructions.  & What is next? / Next is? / What should I do now? \\
          req\_repeat     & User asks for repeating a mentioned entity or instruction. & Can you repeat the first ingredient? / Could you tell me the last step again? \\
          req\_amount     & User asks for the exact quantity of an entity.  & How much sugar do I need? / How much does a package of cheese weigh to? \\
          req\_ingredient & User asks for the next ingredient.  & What is next? / What is next ingredient? \\
          req\_use\_all    & User wants to know if a specific ingredient shall be used up.  & Can I use all the pepper? / Shall I add all the berries? \\
          req\_title      & User asks for the recipe title.   & What are we making today? \\
          req\_is\_recipe\_finished  & User asks if the cooking process ends or not.  & Is it done? / Is the recipe finished? / Is it over? \\
          req\_tool       & User requests a specific tool entity.  & What should I use to make the star shape? \\
          req\_duration   & User requests the time needed for an action.  & How long shall I bake the cake? / When do I know it is ready? \\
          req\_is\_recipe\_ongoing  & User asks if there are more steps to follow.  & Are there more steps? / Anything else to do? \\
          req\_substitute & User wants to know if it is possible to use an alternative ingredient instead of the prescribed one.  & Can I use white sugar instead of brown? \\
          req\_ingr\_list & User requests a detailed list of ingredients before the instruction part starts. & Can you give me all the ingredients I need? \\
          req\_ingr\_list\_length & User requests the total number of ingredients. & How many ingredients do I need? \\
          req\_ingr\_list\_ends  & User wants to know if all the ingredients have been introduced.  & Is it over? / That is all the ingredients? \\
          req\_parallel\_action & User asks for another plausible action while waiting for the current one to be finished. & What can I do in the meantime? \\
          other           & Used for user utterances that cannot be attributed to any of intents above. & - \\
    \end{longtable}
  
    \begin{longtable}{p{3cm} p{4cm} p{4cm}}
      \caption{Annotation scheme for the agent acts.} \label{tab:detailed_agent_acts} \\
        \toprule
          Agent Acts     & Meaning       & Examples \\
      \midrule
    \endfirsthead

      \caption{(continued) Annotation scheme for the agent acts.} \\
      \toprule
          Agent Acts     & Meaning       & Examples \\
      \midrule
    \endhead

      \midrule
        \multicolumn{3}{r}{\footnotesize\itshape Continue on the next page}
    \endfoot

      \bottomrule
    \endlastfoot
        
          greeting        & Agent responds to the user's ``greeting''.  &  Hi. / Greetings. \\
          goodbye         & Agent responds to the user's ``goodbye''.  & Bye. / Have a nice day. \\
          affirm          & Agent gives a positive statement.  &  Yes, you need 6 eggs. / Yes, the recipe is finished. \\
          negate          & Agent gives a negative statement.  & No, you still have to wait. \\
          enjoy           & Agent wishes the user to enjoy the food. &  Enjoy the juicy burger.\\
          end\_recipe      & Agent states that the cooking process is finished.  & This is the last step. \\
          thank           & Agent acknowledges user's gratitude.    &  You are welcome. \\
          fence           & Agent cannot answer the user's question.  & Sorry, I cannot answer that. / Sorry, it is beyond my knowledge.\\ 
          count\_ingredient\_list      & Agent gives the total number of all ingredients.   & We will use 10 ingredients in total. \\
          propose\_start\_recipe & Agent proposes to start the cooking process.  & Shall we start now? / Are you prepared? \\ 
          end\_ingredient\_list & Agent states that introducing ingredients is finished.  & That is all the ingredients we need. / This is the last ingredient. \\
          propose\_list\_ingr     & Agent proposes to introduce ingredient details before the instruction part start. & Do you want to prepare the ingredients beforehand? \\
          propose\_skip\_ing\_list     & Agent proposes to skip the ingredient list if it is too long.  & Do you want me to announce the ingredients when needed? \\
          propose\_next\_inst & Agent proposes to continue to give the next instruction.  & Shall we continue? / Are you prepared for the next step? \\
          propose\_start\_inst    & Agent proposes to start the instruction part after finishing introducing all the ingredients.  & Do you want to know the first step? \\
          propose\_other\_help      & Agent asks the user if he or she needs extra help. (normally this occurs at the end of a conversation.)  & Anything else I can do for you? \\
          inform\_instruction  & Agent informs one instruction symbolized by its step identifier.  & The next step is to remove the cake from oven. \\
          inform\_ingredient   & Agent informs one ingredient symbolized by its step identifier.  & You also need 1 bottle of honey. \\
          inform\_title   & Agent informs the recipe title.  & We will cook Flemish Stew today. \\
          inform\_duration  & Agent informs time duration of an action.  & Are there more steps? / Anything else to do? \\
          inform\_temperature & Agent informs a temperature expression.  & Set the oven to 220 degrees C.\\
          inform\_amount   & Agent informs quantity of an ingredient. & Please add 1/2 teaspoon of salt. \\
          inform\_tool     & Agent informs a tool used in an action. & Prepare 2 feet long lining paper.  \\
          fetch           & Agent asks the user to prepare necessary stuff for one instruction.  & Take a large bowl. \\
          repeat          & Agent repeats a mentioned entity or instruction. It normally appears after the user intent ``req\_repeat''. & - \\
          verify          & Agent checks if the user has the required tool or if the previous instruction is accomplished. & Do you have a slow cooker? / Are both sides of the breast browned?\\
          simplify\_begin & Agent segments a long instruction into a sequence of sub-instructions. This act remarks the first one. & First, add chopped onions.  \\
          simplify\_continue & The other sub-instructions after ``simplify\_begin'' & Then add salad oil and some spices as desired. \\
          other           & Used for agent responses that cannot be attributed to any of acts above. & - \\
    \end{longtable}

\end{appendices}


\bibliography{cook-manuscript-springer}


\end{document}